\documentclass[10pt,twocolumn,letterpaper]{article}

\usepackage[pagenumbers]{cvpr} 

\usepackage{graphicx}
\usepackage{amsmath}
\usepackage{amssymb}
\usepackage{booktabs}
\usepackage{caption}
\usepackage{enumitem}
\usepackage[misc]{ifsym}

\usepackage[accsupp]{axessibility}  

\usepackage[pagebackref,breaklinks,colorlinks]{hyperref}

\usepackage[capitalize]{cleveref}
\crefname{section}{Sec.}{Secs.}
\Crefname{section}{Section}{Sections}
\Crefname{table}{Table}{Tables}
\crefname{table}{Tab.}{Tabs.}

\newcommand{\ys}{\textcolor{black}} 

\def\cvprPaperID{648} 
\def\confName{CVPR}
\def\confYear{2022}

\begin{document}

\title{Unsupervised Image-to-Image Translation with Generative Prior\vspace{-2mm}}

\author{Shuai Yang \hspace{12pt} Liming Jiang  \hspace{12pt}  Ziwei Liu  \hspace{12pt} Chen Change Loy$^{~\textrm{\Letter}}$\\
S-Lab, Nanyang Technological University\\
{\tt\small \{shuai.yang, liming002,  ziwei.liu, ccloy\}@ntu.edu.sg}
}

\twocolumn[{%
\renewcommand\twocolumn[1][]{#1}%
\maketitle
\vspace{-7mm}
\begin{center}
\centering
\includegraphics[width=0.97\linewidth]{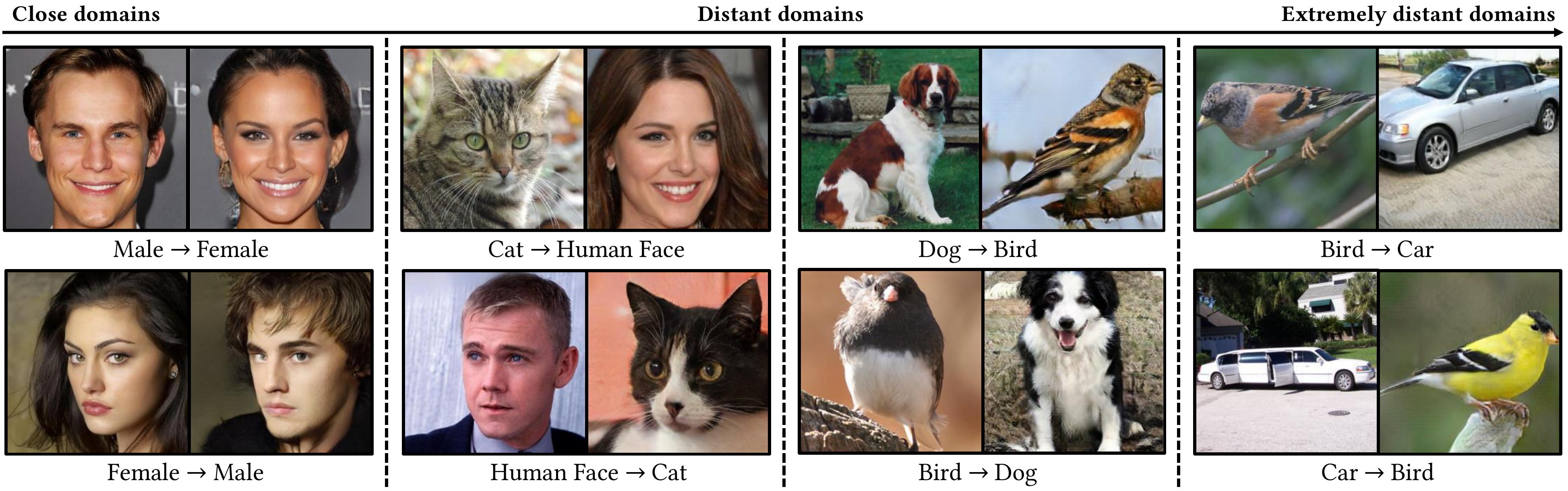}\vspace{-2mm}
\captionof{figure}{We propose a versatile unsupervised image translation framework with generative prior that supports various translations from close domains to distant domains with drastic shape and appearance discrepancies. Each group shows \textit{(left)} the input and \textit{(right)} our result.}
\label{fig:teaser}
\end{center}%
}]

\maketitle

\begin{abstract}
\vspace{-3mm}
  Unsupervised image-to-image translation aims to learn the translation between two visual domains without paired data.
  Despite the recent progress in image translation models, it remains challenging to build mappings between complex domains with drastic visual discrepancies. In this work, we present a novel framework, Generative Prior-guided UNsupervised Image-to-image Translation (\textbf{GP-UNIT}), to improve the overall quality and applicability of the translation algorithm.
  Our key insight is to leverage the generative prior from pre-trained class-conditional GANs (\textit{e.g.}, BigGAN) to learn rich content correspondences across various domains.
  We propose a novel coarse-to-fine scheme:~we first distill the generative prior to capture a robust coarse-level content representation that can link objects at an abstract semantic level, based on which fine-level content features are adaptively learned for more accurate multi-level content correspondences.
  Extensive experiments demonstrate the superiority of our versatile framework over state-of-the-art methods in robust, high-quality and diversified translations, even for challenging and distant domains. Code is available at \url{https://github.com/williamyang1991/GP-UNIT}.\vspace{-5mm}%
\end{abstract}

\section{Introduction}\vspace{-1mm}
\label{sec:intro}

Unsupervised image-to-image translation (UNIT) aims to translate images from one domain to another without paired data.
Mainstream UNIT methods assume a bijection between domains and exploit cycle-consistency~\cite{Zhu2017Unpaired} to build cross-domain mappings. Though good results are achieved in simple cases like horse-to-zebra translations, such assumption is often too restrictive for more general heterogeneous and asymmetric domains in the real world. The performance of existing methods often degrades dramatically in translations with large cross-domain shape and appearance discrepancies such as translating human faces to animal faces, limiting their practical applications.

Translating across domains with large discrepancies requires one to establish the translation at a higher semantic level~\cite{wu2019transgaga}. For instance, to translate a human face to a cat face, one can use the more reliable correspondence of facial components such as the eyes between a human and a cat rather than on the local textures.
In the more extreme case of distant domains, such as animals and man-made objects, a translation is still possible if their correspondence can be determined at a higher abstract semantic level, for example through affirming the frontal orientation of an object or the layout of an object within the image.

Establishing translations at different semantic levels demands a UNIT model's ability to find accurate correspondences of different semantic granularity.~This requirement is clearly too stringent since training a translation model with such a capability requires complex ground truth correspondences that either do not exist or are infeasible to collect.

\begin{figure}[t]
\centering
\includegraphics[width=\linewidth]{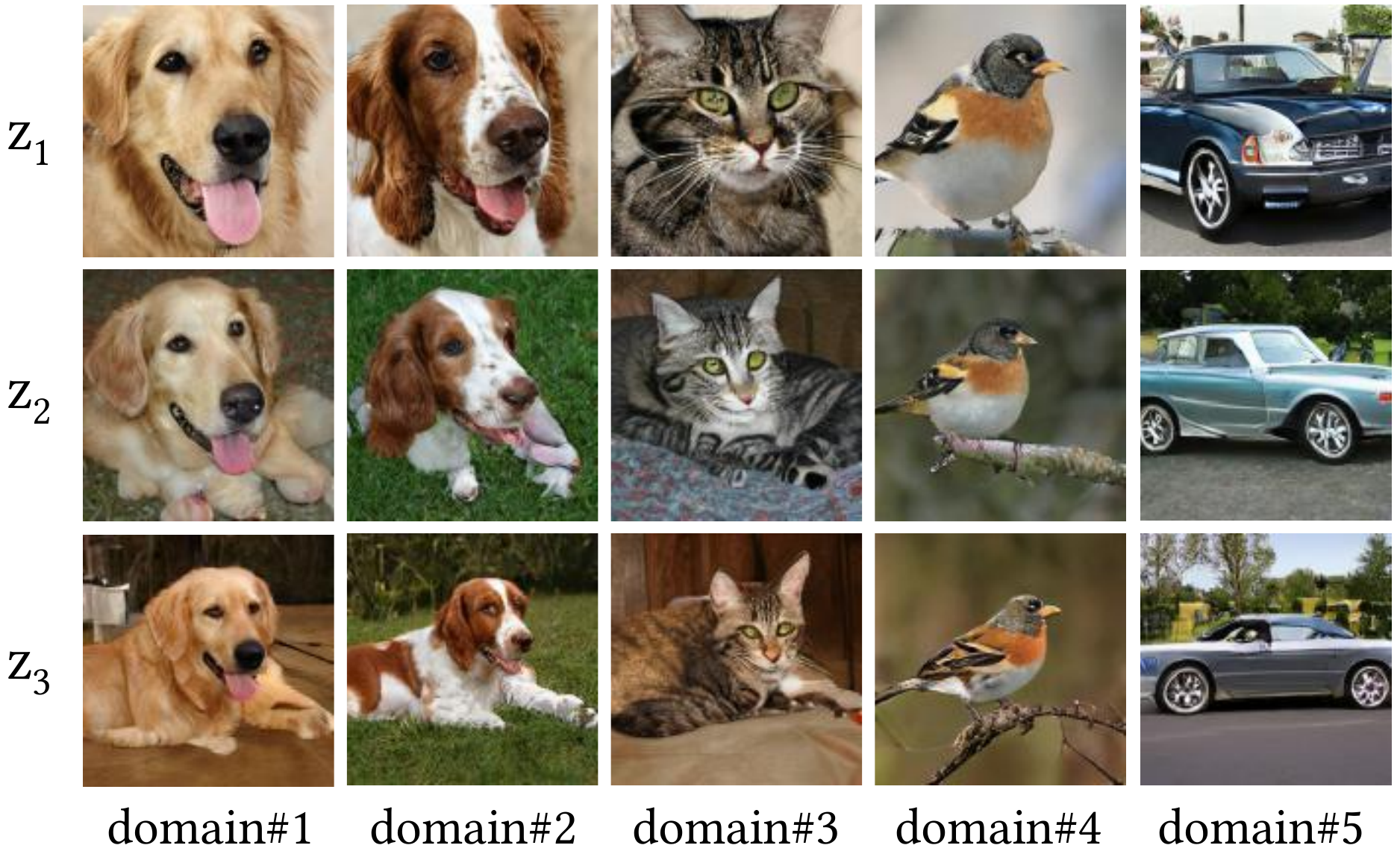}\vspace{-1mm}
\caption{Generative space of BigGAN~\cite{brock2018large}. Objects of different classes generated from the same latent code have a high degree of content correspondences.}\vspace{-5mm}
\label{fig:analysis}
\end{figure}

In this work, we overcome the aforementioned problems through a novel use of generative prior and achieve promising results as shown in Fig.~\ref{fig:teaser}. Specifically, we show that a class-conditional GAN, such as BigGAN~\cite{brock2018large}, provides powerful hints on how different objects are linked -- objects of different classes generated from the same latent code have a high degree of content correspondences (Fig.~\ref{fig:analysis}).
Through generating pairs of such cross-domain images, we can mine the unique prior of the class-conditional GAN and use them to guide an image translation model in building effective and adaptable content mappings across various classes (we will use ``domain'' instead of ``class'' hereafter).

However, such prior is not immediately beneficial to UNIT.
BigGAN, by nature, covers a large number of domains, which makes it an ideal choice of prior for our problem to achieve translation between multiple domains. However, the coverage of many domains inevitably limits the quality and intra-domain diversity of the captured distribution of each domain. Without a careful treatment, such a limitation will severely affect the performance of UNIT in generating high-quality and diverse results.

To overcome the problem above, we decompose a translation task into coarse-to-fine stages: 1) generative prior distillation to learn robust cross-domain correspondences at a high semantic level and 2) adversarial image translation to build finer adaptable correspondences at multiple semantic levels. In the first stage, we train a content encoder to extract disentangled content representation by distilling the prior from the content-correlated data generated by BigGAN.
In the second stage, we apply the pre-trained content encoder to the specific translation task, independent of the generative space of BigGAN,
and propose a dynamic skip connection module to learn adaptable correspondences,
so as to yield plausible and diverse translation results.

To our knowledge, this is the first work to employ BigGAN generative prior for unsupervised\footnote{Following the definition in Liu \etal~\cite{Liu2017Unsupervised}, we call our method unsupervised since our method and the pre-trained BigGAN only use the marginal distributions in individual domains without any explicit cross-domain correspondence supervision.} image-to-image translation.
In particular, we propose a versatile Generative Prior-guided UNsupervised Image-to-image Translation framework (\textbf{GP-UNIT}) to expand the application scenarios of previous UNIT methods that mainly handles close domains.~Our framework shows positive improvements over previous cycle-consistency-guided frameworks in:
1) capturing coarse-level correspondences across various heterogeneous and asymmetric domains, beyond the ability of cycle-consistency guidance;
2) learning fine-level correspondences applicable to various tasks adaptively; and
3) retaining essential content features in the coarse-to-fine stages, avoiding artifacts from the source domain commonly observed in cycle reconstruction.

In summary, our contributions are threefold:
\begin{itemize}[itemsep=1.5pt,topsep=1pt,parsep=0pt]
  \item We propose a versatile GP-UNIT framework that promotes the overall quality and applicability of UNIT with BigGAN generative prior.
  \item We present an effective way of learning robust correspondences across non-trivially distant domains at a high semantic level via generative prior distillation.
  \item We design a novel coarse-to-fine scheme to learn cross-domain correspondences adaptively at different semantic levels.
\end{itemize}

\section{Related Work}
\vspace{-1mm}

\begin{figure*}[t]
\centering
\includegraphics[width=0.93\linewidth]{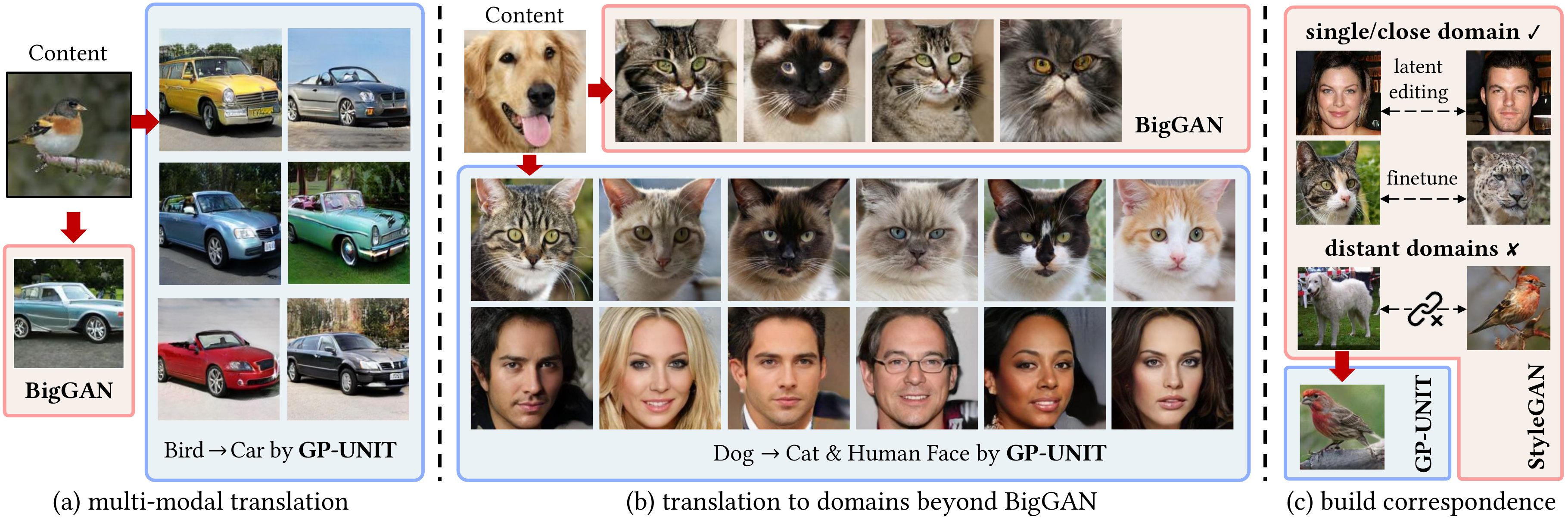}\vspace{-1mm}
\caption{Comparison of the generative spaces of BigGAN, StyleGAN and GP-UNIT. GP-UNIT realizes multi-modal translation, generate cats and human faces beyond ImageNet, and build robust mappings between distant domains. StyleGAN images are from \protect\cite{shen2020interpreting,kwong2021unsupervised}}\vspace{-2mm}
\label{fig:comparison_biggan}
\end{figure*}

\noindent
\textbf{Unsupervised image-to-image translation.}
To learn the mapping between two domains without supervision, CycleGAN~\cite{Zhu2017Unpaired} proposes a novel cycle consistency constraint to build a bi-directional relationship between domains.
To better capture domain-invariant features, representation disentanglement has been investigated extensively in UNIT,
where a content encoder and a style encoder~\cite{Liu2017Unsupervised,Choi2017StarGAN,huang2018multimodal,choi2020stargan,liu2019few,jiang2020tsit} are usually employed to extract domain-invariant content features and domain-specific style features, respectively.
However, learning a disentangled representation between two domains with drastic differences is non-trivial.
To cope with the large visual discrepancy, COCO-FUNIT~\cite{saito2020coco} designs a content-conditioned style encoder to prevent the translation of task-irrelevant appearance information. TGaGa~\cite{wu2019transgaga} uses landmarks to build geometry mappings.
TraVeLGAN~\cite{amodio2019travelgan} proposes a siamese network to seek shared semantic features across domains, and
U-GAT-IT~\cite{kim2019u} leverages an attention module to focus on important regions distinguishing between two domains.
These methods struggle to seek powerful and balanced domain-related representation for specific domains so are less adaptive to the various translation tasks, inevitably failing in certain cases.
Different from these methods, we propose a new coarse-to-fine scheme -- coarse-level cross-domain content correspondences at a highly abstract semantic level are first built, based on which fine-level correspondences adaptive to the task are gradually learned. Such a scheme empowers us to build robust mappings to handle various tasks.

\noindent
\textbf{Adversarial image generation.}
Generative Adversarial Network (GAN)~\cite{goodfellow2014generative} introduces a discriminator to compete with the generator to adversarially approximate the real image distribution.
Among various models, StyleGAN~\cite{karras2019style,karras2020analyzing} has shown promising results.
Many works~\cite{collins2020editing,zhu2020domain,chan2020glean,patashnik2021styleclip,jiang2021talk} exploit the generative prior from StyleGAN to ensure superior image quality by restricting the modulated image to be within the generative space of StyleGAN. However, StyleGAN is an unconditional GAN that is limited to a single domain or close domains~\cite{kwong2021unsupervised}.
BigGAN~\cite{brock2018large} is able to synthesize images in different domains but at the expense of quality and intra-domain diversity. Thus it is not straightforward to exploit BigGAN prior following the aforementioned works.
To circumvent this limitation, in this paper, we distill the generative prior from content-correlated data generated by BigGAN and apply it to the image translation task to generate high-quality images.

\section{Generative Prior Distillation}
\vspace{-1mm}

\subsection{Cross-Domain Correspondences Prior}

Our framework is motivated by the following observation~\cite{AlyafeaiGradient2018Gans,harkonen2020ganspace} -- objects generated by BigGAN, despite originating from different domains, share high content correspondences when generated from the same latent code.
Figure~\ref{fig:analysis} shows the generative space of BigGAN characterized by three latent codes ($z_1$, $z_2$, $z_3$) across five domains.
For each latent code, fine-grained correspondences can be observed between semantically related dogs and cats, such as the face features and body postures.
For birds and vehicles, which are rather different, one can also observe coarse-level correspondences in terms of orientation and layout.

The interesting phenomenon suggests that \textit{there is an inherent content correspondence at a highly abstract semantic level regardless of the domain discrepancy in the BigGAN generative space}.
In particular, objects with the same latent code share either the same or very similar abstract representations in the first few layers, based on which domain-specific details are gradually added.

\begin{figure*}[t]
\centering
\includegraphics[width=0.93\linewidth]{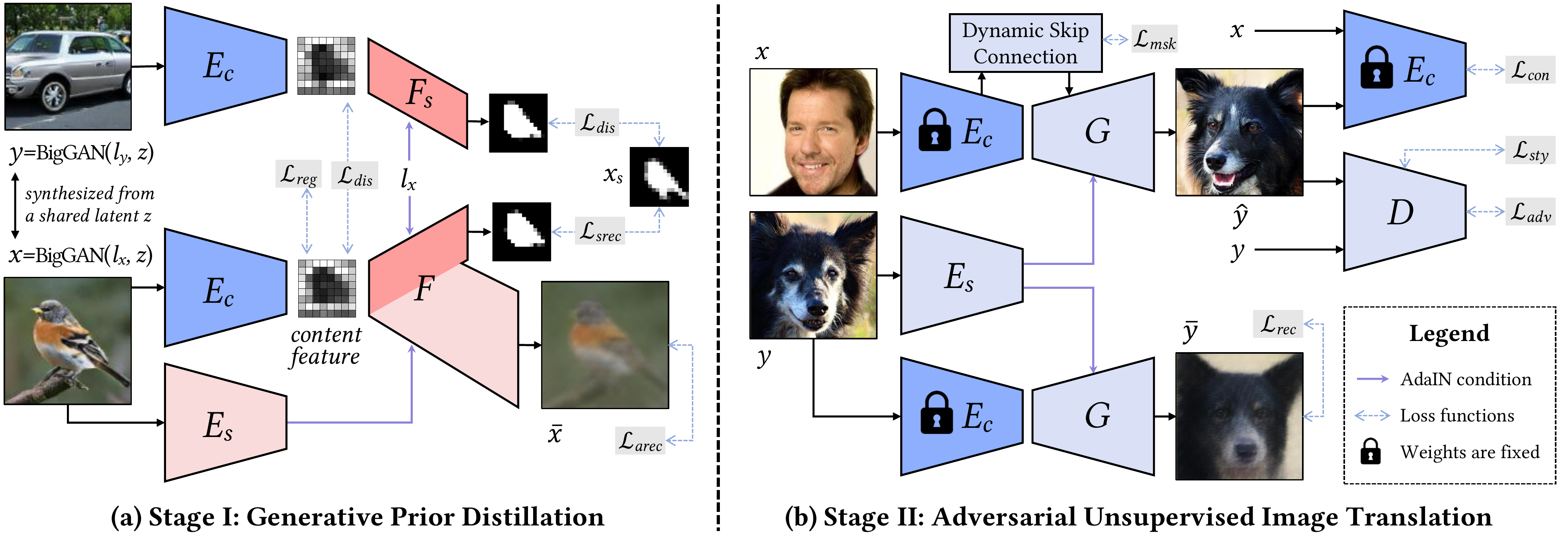}\vspace{-1mm}
\caption{Overview of the proposed GP-UNIT. In the first stage, we use a content encoder $E_c$ to extract shared coarse-level content features between a pair correlated images $(x,y)$ generated by BigGAN from a common random latent code in two random domains. In the second stage, we build our translation network based on the content encoder $E_c$ in the first stage. \ys{For simplicity, we omit the classifier $C$.}}\vspace{-3mm}
\label{fig:framework}
\end{figure*}

In this paper, we exploit this generative prior for building robust mappings and
choose BigGAN for its rich cross-domain prior. Nevertheless, its generative space is limited in \textbf{quality} and \textbf{diversity} for our purpose.
In terms of quality, BigGAN sometimes generates unrealistic objects, such as the dog body with $z_2$ in Fig.~\ref{fig:analysis}.
As for diversity, first, the space lacks intra-domain variation, \eg, the diversity in textures of a dog or colors of a bird in the same domain are pretty limited. Using such prior in UNIT will overfit the model to limited appearances.
Second, the BigGAN generative space is limited to 1,000 domains of ImageNet~\cite{russakovsky2015imagenet}, which is inadequate for an actual UNIT. For instance, it only has four kinds of domestic cats as in Fig.~\ref{fig:comparison_biggan}(b) and excludes the important domain of human faces.

StyleGAN is not suitable for our task despite its generative space is of high quality and diversity. This is because it is limited to a single domain, and thus it is mainly applied to attribute transfer within one domain via latent editing~\cite{collins2020editing,zhu2020domain,patashnik2021styleclip}.
Recently, cross-domain translations on StyleGAN have been achieved via finetuning~\cite{pinkney2020resolution,kwong2021unsupervised}, but this still assumes a small distance between the models of the source and target domains and is therefore still restricted to close domains.
The assumption makes StyleGAN prior less applicable to more complex translation tasks.

Our framework solves the above problems by distilling a general generative prior from BigGAN instead of directly constraining the latent or image space. It enables us to design and train the translation module independently. Therefore, we can realize multi-modal translation (Fig.~\ref{fig:comparison_biggan}(a)), generalize to classes beyond ImageNet (Fig.~\ref{fig:comparison_biggan}(b)) and build robust mappings between distant domains (Fig.~\ref{fig:comparison_biggan}(c)). Next, we detail how we distill the prior.

\subsection{Prior Distillation with Content Encoder}

Given the correlated images $(x,y)$ generated by BigGAN from a common random latent code in two random domains $\mathcal{X}$ and $\mathcal{Y}$, our main goal is to train a content encoder $E_c$ to extract their shared coarse-level content features, which can be used to reconstruct their shape and appearance.
Figure~\ref{fig:framework}(a) illustrates this autoencoder pipeline for generative prior distillation.

Specifically, we use a decoder $F$ to recover the appearance $x$ based on its content feature $E_c(x)$, style feature $E_s(x)$ extracted by a style encoder $E_s$ and the domain label $l_x$.
We further exploit the shallow layers $F_s$ of $F$ to predict the shape of $x$ (\textit{i.e.},  instance segmentation map $x_s$, which is extracted from $x$ by HTC~\cite{chen2019hybrid}) based on $E_c(x)$ and $l_x$. We find such auxiliary prediction eases the training on hundreds of domains. Besides the shape and appearance reconstruction, we further regularize the content feature in three ways for disentanglement: 1) $x$ and $y$ should share the same content feature; 2) We introduce a classifier $C$ with a gradient reversal layer $R$~\cite{ganin2015unsupervised} to make the content feature domain-agnostic; 3) We limit $E_c(x)$ to one channel to eliminate domain information~\cite{sushko2021one} and add Gaussian noise of a fixed variance for robustness. Our objective function is:
\begin{equation}\label{eq:total_loss2}
  \min_{E_c,E_s,F,C}\mathcal{L}_{arec}+\mathcal{L}_{srec}+\mathcal{L}_{dis}+\mathcal{L}_{reg},
\end{equation}
where $\mathcal{L}_{arec}$ is the appearance reconstruction loss measuring the $L_2$ and perceptual loss~\cite{Johnson2016Perceptual} between $\bar{x}=F(E_c(x),E_s(x),l_x)$ and $x$. The shape reconstruction loss, $\mathcal{L}_{srec}$, is defined as
\begin{equation}
  \mathcal{L}_{srec}=\lambda_{s}\mathbb{E}_{x}[\|F_s(E_c(x),l_x)-x_s\|_1].
\end{equation}
The binary loss $\mathcal{L}_{dis}$ with paired inputs narrows the distance between the content features of $x$ and $y$. In addition, we would like to recover the shape of $x$ with the content feature of $y$, which simulates translations:
\begin{equation}
  \mathcal{L}_{dis}=\mathbb{E}_{(x,y)}[\|E_c(x)-E_c(y)\|_1+\lambda_{s}\|F_s(E_c(y),l_x)-x_s\|_1].  \nonumber
\end{equation}
Finally, $\mathcal{L}_{reg}$ guides $C$ to maximize the classification accuracy and pushes $E_c$ to confuse $C$, so that the content feature is domain-agnostic.
An $L_2$ norm is further applied to the content feature:
\begin{equation}
  \mathcal{L}_{reg}=\mathbb{E}_{x}[-l_x\log C(R(E_c(x)))]+\lambda_{r}\mathbb{E}_{x}[\|E_c(x)\|_2].  \nonumber
\end{equation}
For unary losses of $\mathcal{L}_{arec}$, $\mathcal{L}_{srec}$ and $\mathcal{L}_{reg}$, we also use real images of ImageNet~\cite{russakovsky2015imagenet} and CelebA-HQ~\cite{karras2018progressive} for training to make $E_c$ more generalizable.

\section{Adversarial Image Translation}

Given a fixed content encoder $E_c$ pre-trained in the first stage, we build our translation network following a standard style transfer paradigm in the second stage. Thanks to the pre-trained $E_c$ that provides a good measurement for content similarity, our framework does not need cycle training.

As shown in Fig.~\ref{fig:framework}(b), our translation network receives a content input $x\in\mathcal{X}$ and a style input $y\in\mathcal{Y}$. The network extracts their content feature $E_c(x)$ and style feature $E_s(y)$, respectively. Then, a generator $G$ modulates $E_c(x)$  to match the style of $y$ via AdaIN~\cite{huang2017adain}, and finally produces the translated result $\hat{y}=G(E_c(x),E_s(y))$. The realism of $\hat{y}$ is reinforced through an adversarial training with a discriminator $D$,
\begin{equation}
  \mathcal{L}_{adv}=\mathbb{E}_{y}[\log D(y)]+\mathbb{E}_{x,y}[\log (1-D(\hat{y}))].
\end{equation}
In addition, $\hat{y}$ is required to fit the style of $y$, while preserving the original content feature of $x$, which can be formulated as a style loss $\mathcal{L}_{sty}$ and a content loss $\mathcal{L}_{con}$,
\begin{align}
  \mathcal{L}_{sty}&=\mathbb{E}_{x,y}[\|f_D(\hat{y})-f_D(y)\|_1],\\
  \mathcal{L}_{con}&=\mathbb{E}_{x,y}[\|E_c(\hat{y})-E_c(x)\|_1],
\end{align}
where $f_D$ is the style feature defined as the channel-wise mean of the middle layer feature of $D$ following the style definition in~\cite{huang2017adain}.

\subsection{Dynamic Skip Connection}

Domains that are close semantically would usually exhibit fine-level content correspondences that cannot be characterized solely by the abstract content feature. To solve this problem, we propose a dynamic skip connection module, which passes middle layer features $f_E$ from $E_c$ to $G$ and predicts masks $m$ to select the valid elements for establishing fine-level content correspondences.

Our dynamic skip connection is inspired by the GRU-like selective transfer unit~\cite{liu2019stgan}.
Let the superscript $l$ denote the layer of $G$.
The mask $m^l$ at layer $l$ is determined by the encoder feature $f_E^l$ passed to the same layer and a hidden state $h^{l-1}$ at the last layer.
Specifically, we first set the first hidden state $h^0=E_c(x)$ and use the upsampling convolution to match the dimension of $h^{l-1}$ with $f_E^l$ as $\hat{h}^{l-1}=\sigma(W^l_h\circ \uparrow h^{l-1})$, where $\uparrow$, $\circ$ and $W^l_h$ are the upsample operator, convolution operator and convolution weights, respectively. The activation layer is denoted as $\sigma$. Then, our module at layer $l$ updates the hidden state $h^l$ and the encoder feature $\hat{f}_E^l$, and fuses $\hat{f}_E^l$ with the generator feature $f_G^l$ with the predicted mask $m^l$:
\begin{align}
  r^l=\sigma(W^l_r\circ [\hat{h}^{l-1},f_E^l]),&~~~
  m^l=\sigma(W^l_m\circ [\hat{h}^{l-1},f_E^l]),   \nonumber\\
  h^l=r^l\hat{h}^{l-1},&~~~
  \hat{f}_E^l=\sigma(W^l_E\circ [h^l,f_E^l]),   \nonumber\\
  f^l=(1-&m^l)f_G^l+m^l\hat{f}_E^l,   \nonumber
\end{align}
where $[\cdot,\cdot]$ denotes concatenation. $m^l$ has the same dimension of $f_G^l$, serving both channel attention and spatial attention.
Moreover, we apply $L_1$ norm to $m^l$ to make it sparse,
\begin{equation}
  \mathcal{L}_{msk}=\sum\nolimits_l\mathbb{E}_{x}[\|m^l\|_1],
\end{equation}
so that only the most useful content cues from the source domain are selected.

\noindent
\textbf{Full objectives.}
Combining the aforementioned losses, our full objectives take the following form:
\begin{equation}\label{eq:total_loss}
  \min_{G,E_s}\max_{D}\mathcal{L}_{adv}+\lambda_{1}\mathcal{L}_{con}+\lambda_{2}\mathcal{L}_{sty}
  +\lambda_{3}\mathcal{L}_{msk}+\lambda_{4}\mathcal{L}_{rec}. \nonumber
\end{equation}
A reconstruction loss $\mathcal{L}_{rec}$ is added to measure the $L_1$ and perceptual loss~\cite{Johnson2016Perceptual} between $y$ and $\bar{y}=G(E_c(y),E_s(y))$.
Intuitively, we would like the learned style feature of an image to precisely reconstruct itself with the help from its content feature, which stabilizes the network training.

\noindent
\textbf{Style sampling.}
To sample latent style features directly for multi-modal generation without the style images, we follow the post-processing of~\cite{meshry2021step} to train a mapping network to map the unit Gaussian noise to the latent style distribution using a maximum likelihood
criterion~\cite{hoshen2019non}. Please refer to~\cite{hoshen2019non} for the details.

\section{Experimental Results}
\label{sec:experiment}

\begin{figure*}[t]
\centering
\includegraphics[width=1\linewidth]{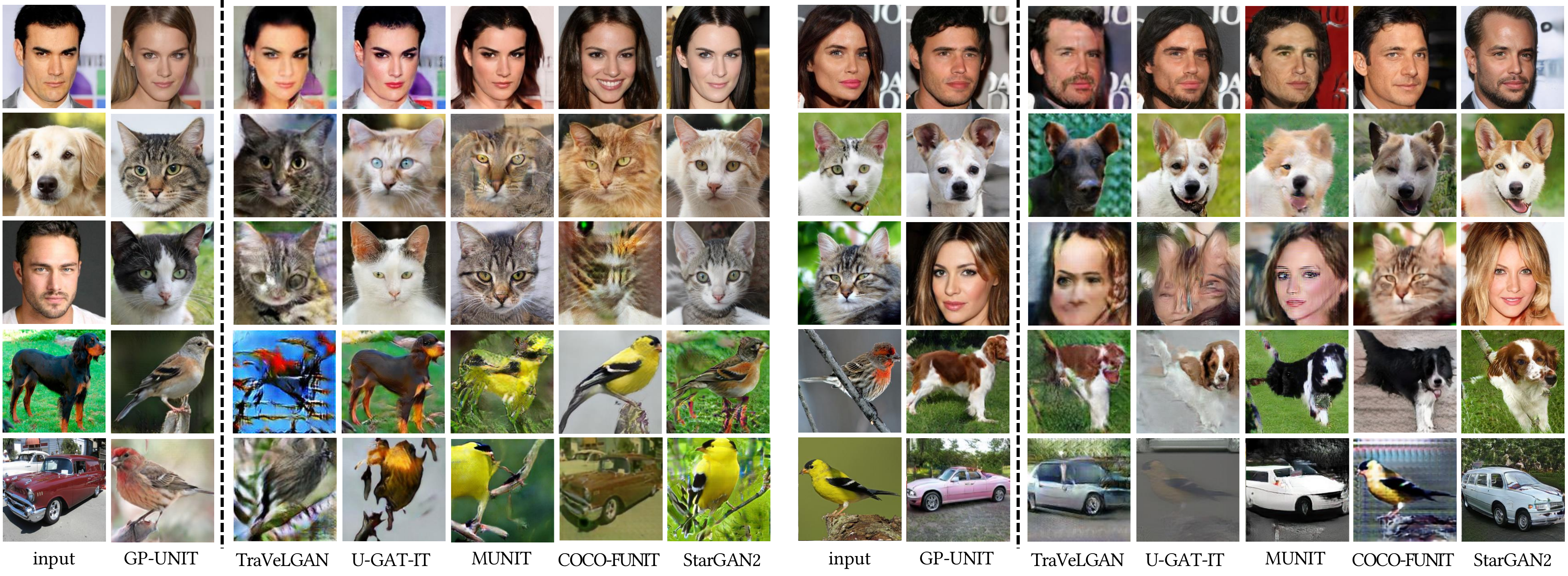}\vspace{-1mm}
\caption{Visual comparison with TraVeLGAN~\cite{amodio2019travelgan}, U-GAT-IT~\cite{kim2019u}, MUNIT~\cite{huang2018multimodal}, COCO-FUNIT~\cite{saito2020coco} and StarGAN2~\cite{choi2020stargan}. GP-UNIT consistently outperforms on all tasks and demonstrates greater superiority as the task becomes more challenging (from top to bottom).}
\vspace{-2mm}
\label{fig:comparison}
\end{figure*}

\noindent
\textbf{Dataset.} In the first stage, we prepare both synthesized data and real data.
For synthesized data, we use the official BigGAN
\cite{brock2018large} to generate correlated images associated by random latent codes for each of the 291 domains including animals and vehicles. After filtering low-quality ones, we finally obtain 655 images per domain that are linked across all domains, 600 of which are for training. We denote this dataset as \textit{synImageNet-291}.
For real data, we apply HTC~\cite{chen2019hybrid} to ImageNet~\cite{russakovsky2015imagenet} to detect and crop the object regions. Each domain uses 600 images for training. We denote this dataset as \textit{ImageNet-291}.
Besides, 29K face images of CelebA-HQ~\cite{liu2015deep, karras2018progressive} are also included for training.

In the second stage, we perform evaluation on four translation tasks.~1) Male $\leftrightarrow$ Female on 28K training images of  CelebA-HQ~\cite{liu2015deep, karras2018progressive}.
2) Dog $\leftrightarrow$ Cat on AFHQ~\cite{choi2020stargan}, with 4K training images per domain.
3) Human Face $\leftrightarrow$ Cat on 4K AFHQ images and 29K CelebA-HQ images.
4) Bird $\leftrightarrow$ Dog or Car:
Four classes of birds, four classes of dogs
and four classes of cars
in \textit{ImageNet-291} are used. Every four classes form a domain with 2.4K training images.
\ys{Here, we use Bird$\leftrightarrow$ Car as the extreme case to test to what extent GP-UNIT can handle for stress testing.}

\noindent
\textbf{Network training.} We set $\lambda_{s}=5$, $\lambda_{r}=0.001$, $\lambda_{1}=\lambda_{3}=\lambda_{4}=1$ and $\lambda_{2}=50$.
For Cat $\rightarrow$ Human Face, we use an additional identity loss~\cite{deng2019arcface} with weight $1$ to preserve the identity of the reference face following~\cite{richardson2020encoding}.
Dynamic skip connections are applied to the 2nd layer ($l=1$) and the 3rd layer ($l=2$) of $G$.
Except for Male $\leftrightarrow$ Female, we do not use dynamic skip connections to compute $\mathcal{L}_{rec}$ (by setting $m^l$ to an all-zero tensor), which are discussed in Sec.~\ref{sec:more_result}.

\subsection{Comparison with the State of the Arts}

\begin{figure}[t]
\centering
\includegraphics[width=1\linewidth]{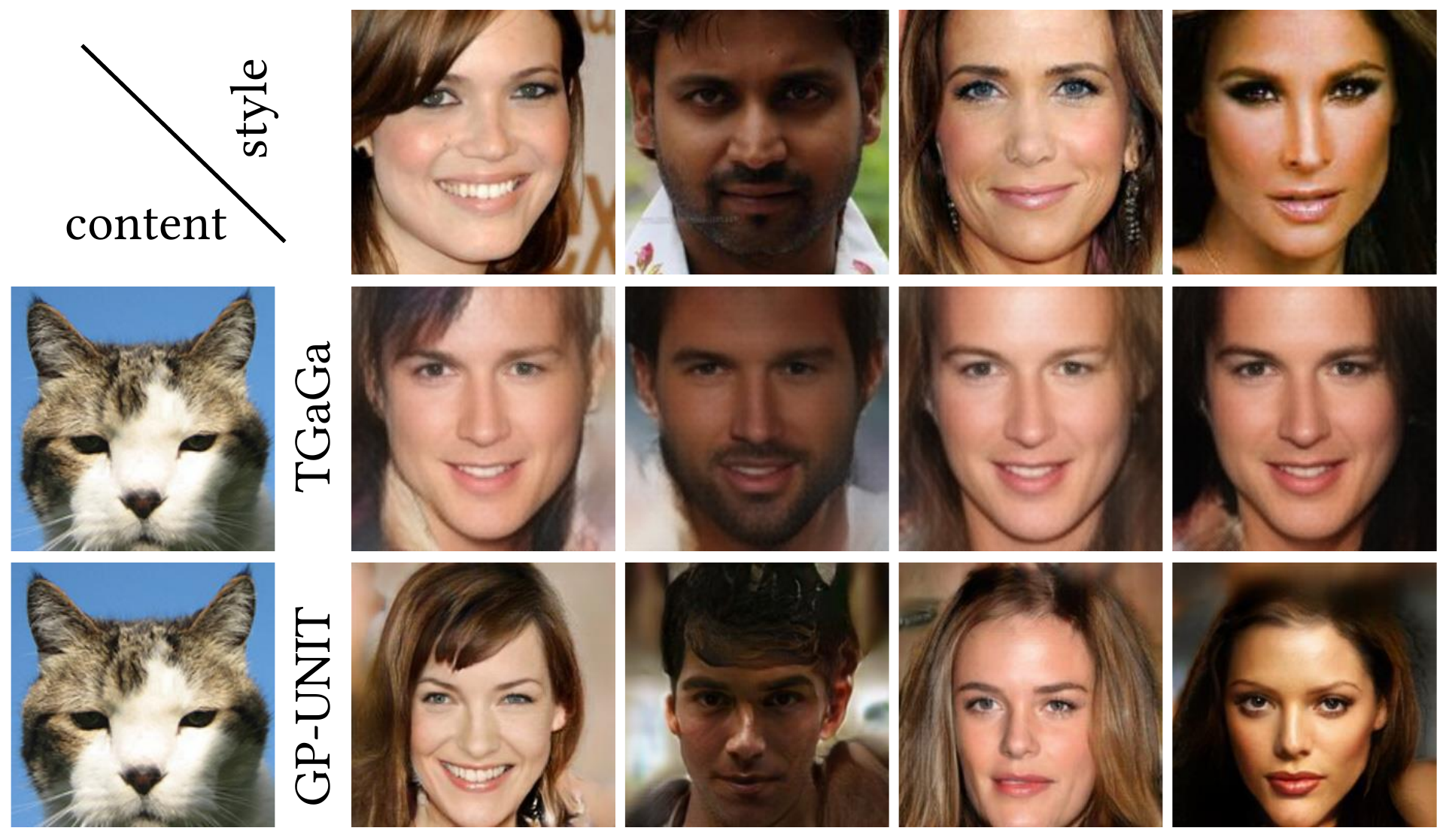}\vspace{-2mm}
\caption{Comparing exemplar-guided translation with TGaGa\protect\footnotemark. GP-UNIT surpasses TGaGa in vivid details and style consistency.}\vspace{-4mm}
\label{fig:transgaga}
\end{figure}

\noindent
\textbf{Qualitative comparison.} We perform visual comparison to six state-of-the-art methods in Fig.~\ref{fig:comparison} and Fig.~\ref{fig:transgaga}.
As shown in Fig.~\ref{fig:comparison}, cycle-consistency-guided U-GAT-IT~\cite{kim2019u}, MUNIT~\cite{huang2018multimodal} and StarGAN2~\cite{choi2020stargan} rely on the low-level cues of the input image for bi-directional reconstruction, which leads to some undesired artifacts, such as the distorted cat face region that corresponds to the dog ears, and the ghosting dog legs in the generated bird images.
Meanwhile, TraVeLGAN~\cite{amodio2019travelgan} and COCO-FUNIT~\cite{saito2020coco} fail to build proper content correspondences for Human Face $\leftrightarrow$ Cat and Bird $\leftrightarrow$ Car.~By comparison, our method is comparable to the above methods on Male $\leftrightarrow$ Female task and show consistent superiority on other challenging tasks.
In Fig.~\ref{fig:transgaga}, we compare our model to TGaGa~\cite{wu2019transgaga}, which also deals with large geometric deformations
on exemplar-guided translation.
TGaGa produces blurry results and fails to match the example appearance precisely, \textit{e.g.}, all the generated faces look alike, except for the color changes. %
GP-UNIT surpasses TGaGa in both vivid details and style consistency.

\begin{table*} [t]
\caption{Quantitative comparison. We use FID and Diversity with LPIPS to evaluate the quality and diversity of the generated images.}\vspace{-1mm}
\label{tb:fid}
\centering
\scriptsize
\begin{tabular*}{\textwidth}{@{\extracolsep{\fill}}l|c|c|c|c|c|c|c|c|c|c|c|c}
\toprule
Task & \multicolumn{2}{c|}{Male $\leftrightarrow$ Female} &  \multicolumn{2}{c|}{Dog $\leftrightarrow$ Cat} & \multicolumn{2}{c|}{Human Face $\leftrightarrow$ Cat}  &  \multicolumn{2}{c|}{Bird $\leftrightarrow$ Dog} &  \multicolumn{2}{c|}{Bird $\leftrightarrow$ Car} &   \multicolumn{2}{c}{Average} \\
\midrule
Metric & FID & Diversity & FID & Diversity & FID & Diversity & FID & Diversity & FID & Diversity & FID & Diversity \\
\midrule
TraVeLGAN & 66.60  & $-$ & 58.91  & $-$ & 85.28  & $-$ & 169.98  & $-$ & 164.28  & $-$ & 109.01  & $-$ \\
U-GAT-IT & 29.47  & $-$ & 38.31  & $-$ & 110.57  & $-$ & 178.23  & $-$ & 194.05  & $-$ & 110.12  & $-$ \\
\midrule
MUNIT & 22.64  & 0.37  & 80.93  & 0.47  & 56.89  & \textbf{0.53}  & 217.68  & 0.57  & 121.02  & 0.60  & 99.83  & 0.51 \\
COCO-FUNIT & 39.19  & 0.35  & 97.08  & 0.08  & 236.90  & 0.33  & 30.27  & 0.51  & 207.92  & 0.12  & 122.27  & 0.28 \\
StarGAN2 & \textbf{14.61}  & \textbf{0.45}  & 22.08  & 0.45  & \textbf{11.35}  & 0.51  & 20.54  & 0.52  & 29.28  & 0.58  & 19.57  & 0.50 \\
\textbf{GP-UNIT} & 14.63  & 0.37  & \textbf{15.29}  & \textbf{0.51}  & 13.04  & 0.49  & \textbf{11.29}  & \textbf{0.60}  & \textbf{13.93}  & \textbf{0.61}  & \textbf{13.64}  & \textbf{0.52} \\
\bottomrule
\end{tabular*}\vspace{-3mm}
\end{table*}

\footnotetext{At the time of this submission, the code and training data of TGaGa are not released. We directly use the test and result images kindly provided by the authors of TGaGa. Since the training data of GP-UNIT and TGaGa does not match, this comparison is for visual reference only.}

\begin{figure*}[t]
\centering
\includegraphics[width=0.98\linewidth]{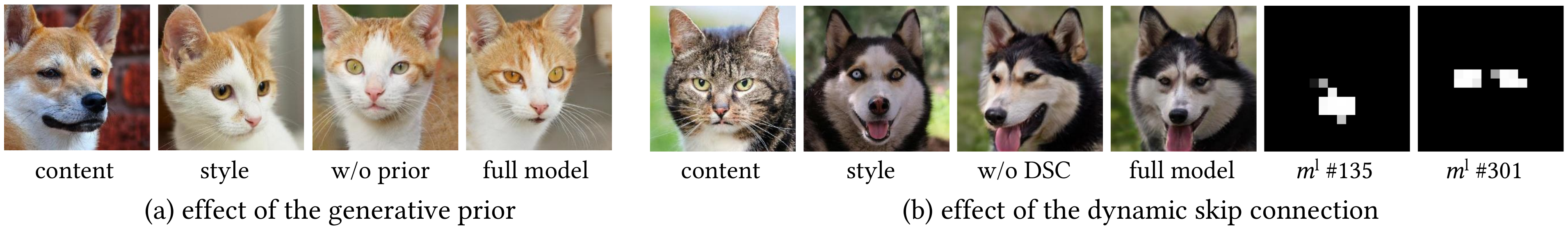}\vspace{-3mm}
\caption{Ablation study on the generative prior and the dynamic skip connection.}\vspace{-2mm}
\label{fig:ablation}
\end{figure*}
\begin{figure*}[t]
\centering
\includegraphics[width=0.98\linewidth]{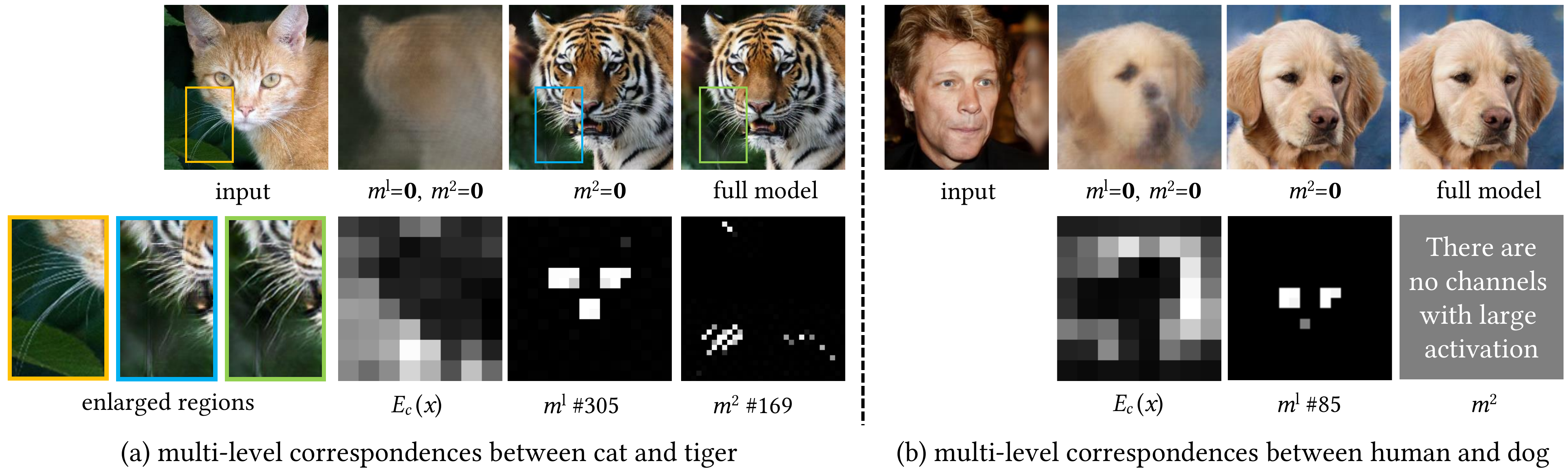}\vspace{-2mm}
\caption{Our framework learns multi-level content correspondences that are robust and adaptable to different translation tasks.}\vspace{-3mm}
\label{fig:correspondence}
\end{figure*}

\begin{table} [t]
\caption{User preference scores in terms of content consistency and overall preference. Best scores are marked in bold.}\vspace{-2mm}
\label{tb:user_study}
\centering
\scriptsize
\begin{tabular}{l|c|c}
\toprule
Metric & Content Consistency & Overall Preference \\
\midrule
TraVeLGAN & 0.012 & 0.006 \\
U-GAT-IT  & 0.076  & 0.050 \\
MUNIT  &  0.045  & 0.033 \\
COCO-FUNIT &  0.065  & 0.044 \\
StarGAN2 &  0.199  & 0.171 \\
\textbf{GP-UNIT} & \textbf{0.603} & \textbf{0.696} \\
\bottomrule
\end{tabular}\vspace{-4mm}
\end{table}

\noindent
\textbf{Quantitative comparison.}~We follow~\cite{wu2019transgaga,choi2020stargan} to perform quantitative comparison  in terms of quality and diversity.
FID~\cite{heusel2017gans} and LPIPS~\cite{zhang2018unreasonable} are used to evaluate the image quality of generated results against the real data and the output diversity, respectively.
For methods supporting multi-modal transfer (MUNIT, COCO-FUNIT, StarGAN2, GP-UNIT), we generate 10 paired translation results per test image from randomly sampled latent codes or exemplar images to compute their average LPIPS distance.
The quantitative results averaged over all test images are reported in Table~\ref{tb:fid}, which
are consistent with Fig.~\ref{fig:comparison}, \ie, our method is comparable or superior to the compared methods, and the advantage becomes more distinct on difficult tasks, obtaining the best overall FID and diversity.
We find GP-UNIT tends to preserve the background of the input image. This property does not favor diversity, but might be useful in some applications.
Although StarGAN2 yields realistic human faces  (best FID) on Cat $\rightarrow$ Human Face, it ignores the pose correspondences with the input cat faces (lower content consistency than GP-UNIT), as in Fig.~\ref{fig:comparison}.

We further conduct a user study to evaluate the input-output content consistency and overall translation performance. A total of 25 subjects participate in this study to select what they consider to be the best results from the six methods, and a total of 2,500 selections on 50 groups of results are tallied. Table~\ref{tb:user_study} summarizes the average preference scores, where the proposed method receives notable preference for both content consistency and overall performance.

\subsection{Ablation Study}
\label{sec:ablation}

\noindent
\textbf{Generative prior distillation.} As shown in Fig.~\ref{fig:ablation}(a), if we train our content encoder from scratch along with all other subnetworks in the second stage, like most image translation frameworks, this variant fails to preserve the content features such as the eye position. By comparison, our pre-trained content encoder successfully exploits the generative prior to build effective content mappings. It also suggests the necessity of the coarse-level content feature, only based on which valid finer-level features can be learned. Hence, the generative prior is the key to the success of our coarse-to-fine scheme of content correspondence learning.

\noindent
\textbf{Dynamic skip connection.} As shown in Fig.~\ref{fig:ablation}(b), without dynamic skip connections (DSC), the model cannot keep the relative position of the nose and eyes as in the content images. We show that the 135th and 301st channels of the mask $m^1$ predicted by our full model effectively locate these features for accurate content correspondences.

\noindent
\textbf{Multi-level cross-domain correspondences.}~Figure~\ref{fig:correspondence} analyzes the learned multi-level correspondences. The most abstract $E_c(x)$ only gives layout cues. If we solely use $E_c(x)$ (by setting both masks $m^1$ and $m^2$ to all-zero tensors), the resulting tiger and dog faces have no details. Meanwhile, $m^1$ focuses on mid-level details like the nose and eyes of cat face in the 305th channels, and eyes of human face in the 85the channels, which is enough to generate a realistic result with $E_c(x)$. Finally, $m^2$ pays attention to subtle details like the cat whiskers in the 169th channel for close domains.
Therefore, our full multi-level content features enable us to simulate the extremely fine-level long whiskers in the input. As expected, such kind of fine-level correspondences are not found between more distant human and dog faces, preventing the unwanted appearance influence from the source domain (\eg, clothes in the generated cat faces in Fig.~\ref{fig:comparison}).
Note that such reasonable and adaptable semantic attentions are learned merely via the generation prior, without any explicit correspondence supervision.

\begin{figure}[t]
\centering
    \includegraphics[width=0.96\linewidth]{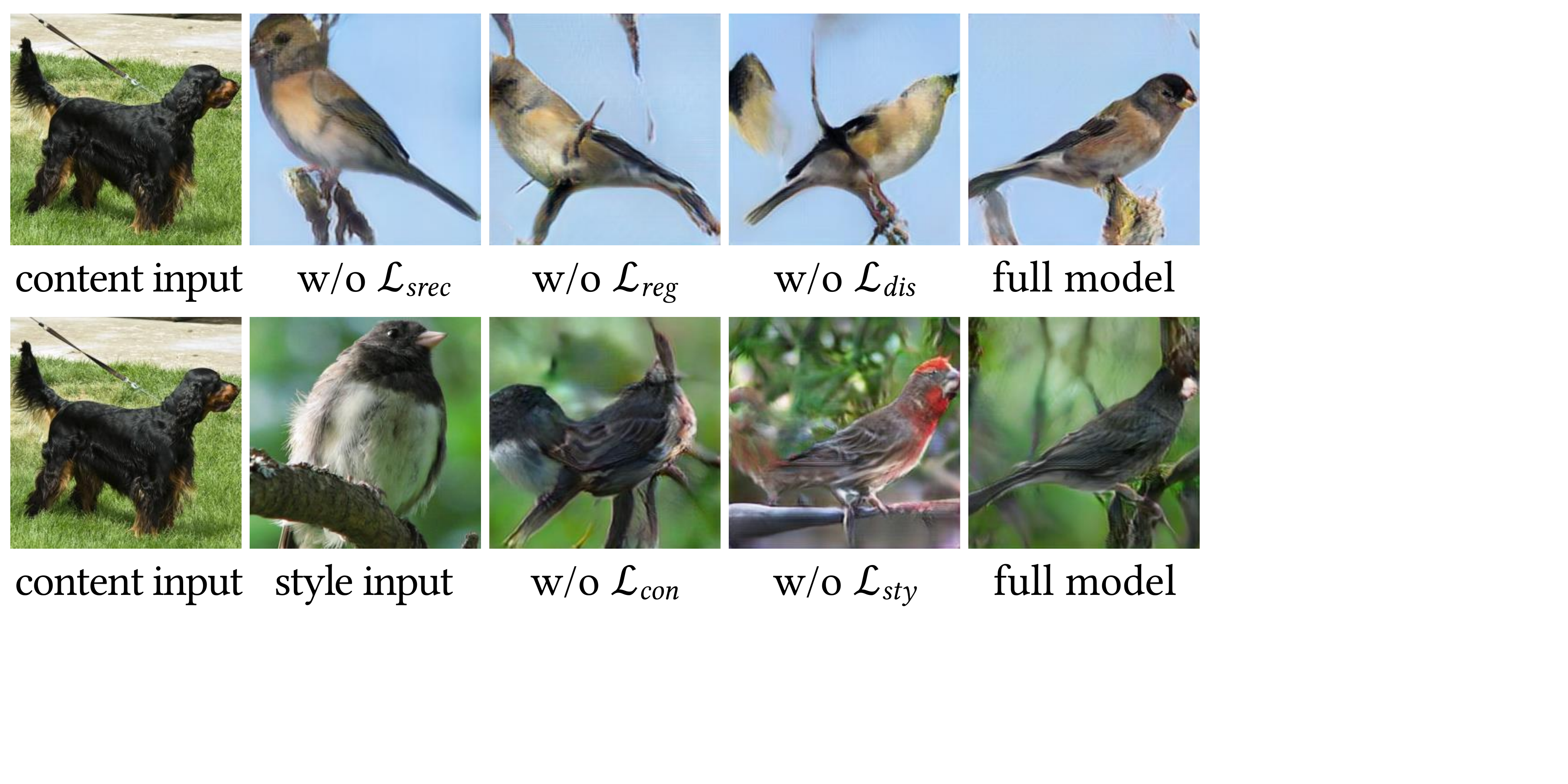}\vspace{-2mm}
\caption{Effects of the loss terms.}\vspace{-5mm}
\label{fig:ablation0}
\end{figure}

\noindent
\ys{\textbf{Loss functions.} Figure~\ref{fig:ablation0} studies the effects of the loss terms.
In Stage I, $\mathcal{L}_{srec}$ is the key to learn correct content features, or correspondence is not built.
$\mathcal{L}_{reg}$ makes content features more sparse to improve robustness to unimportant domain-specific details.
$\mathcal{L}_{dis}$ finds domain-shared features to prevent the output from affected by objects from the source domain like the dog tail.
In Stage II, $\mathcal{L}_{con}$ helps strengthen the pose correspondence while $\mathcal{L}_{sty}$ makes the output better match the style of the  exemplar image.
}

\subsection{More Results}
\label{sec:more_result}

\begin{figure}[t]
\centering
\includegraphics[width=1\linewidth]{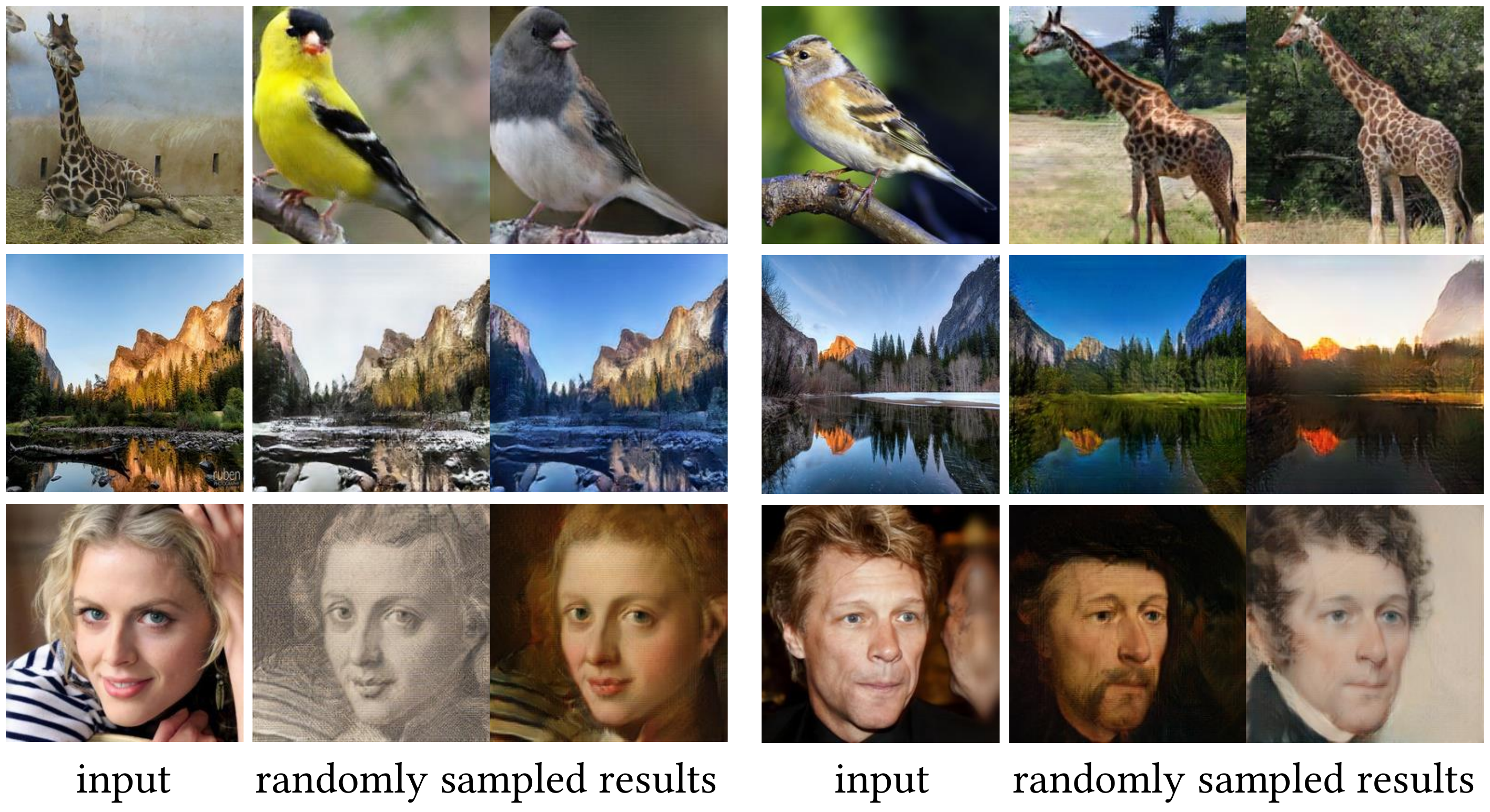}\vspace{-2mm}
\caption{Applicability to domains beyond BigGAN: (\textit{top}) Giraffe $\leftrightarrow$ Bird,  (\textit{middle}) Summer $\leftrightarrow$ Winter, (\textit{bottom}) Face $\rightarrow$ Art.}\vspace{-1mm}
\label{fig:generalization}
\end{figure}

\noindent
\textbf{Generalization to domains beyond BigGAN.}
Figure~\ref{fig:generalization} shows three applications of species translation, season transfer and facial stylization.~Even if MS-COCO giraffes~\cite{lin2014microsoft}, Yosemite landscapes~\cite{huang2018multimodal} and Art portraits~\cite{karras2020training} are not within the ImageNet 1,000 classes and are not observed by the content encoder in Stage I, our method can well support these domains and generate realistic results.

\begin{figure}[t]
\centering
    \includegraphics[width=1\linewidth]{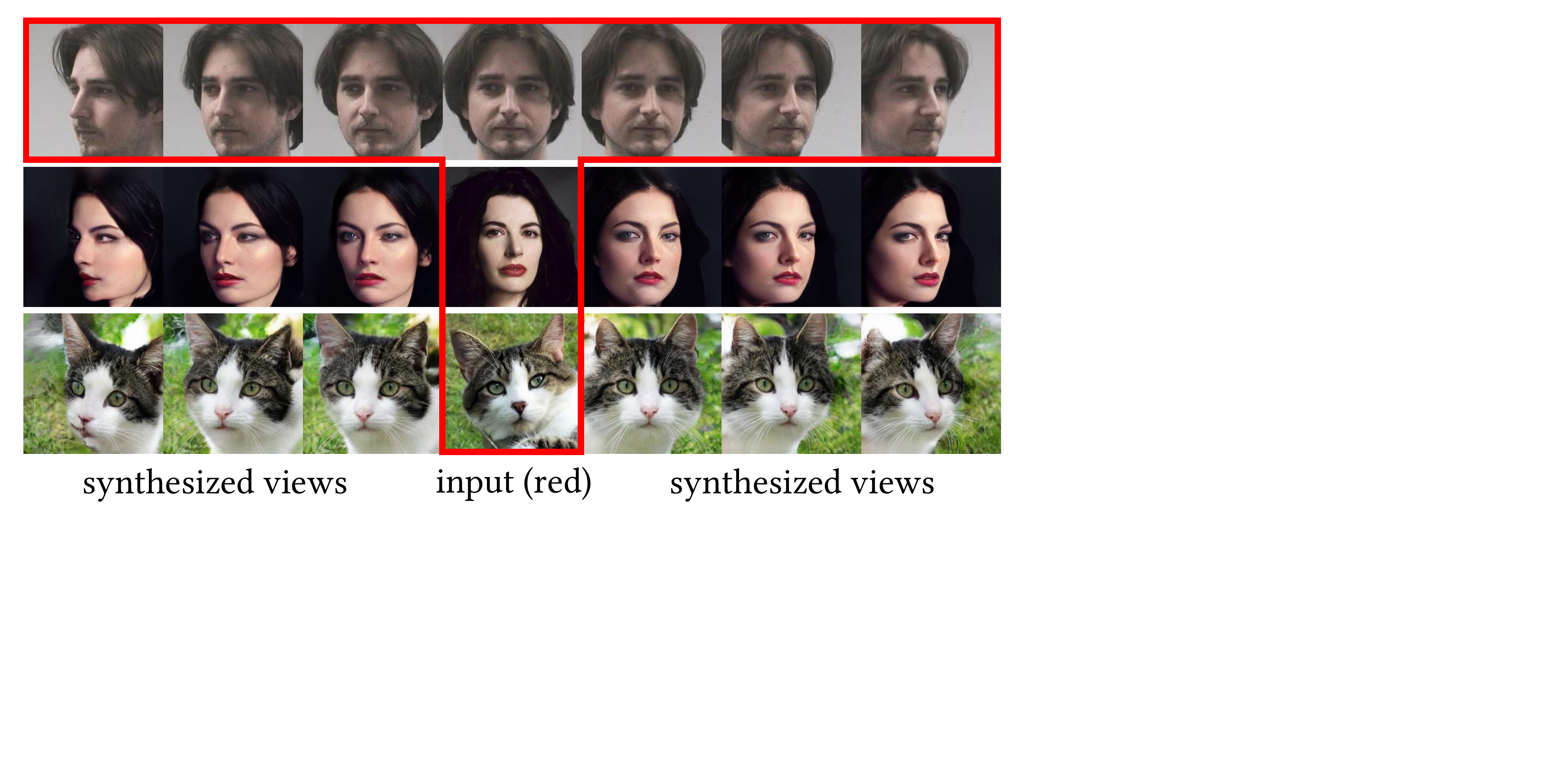}\vspace{-2mm}
\caption{Unseen view synthesis.}\vspace{-2mm}
\label{fig:interpolation2}
\end{figure}

\begin{figure}[t]
\centering
    \includegraphics[width=1\linewidth]{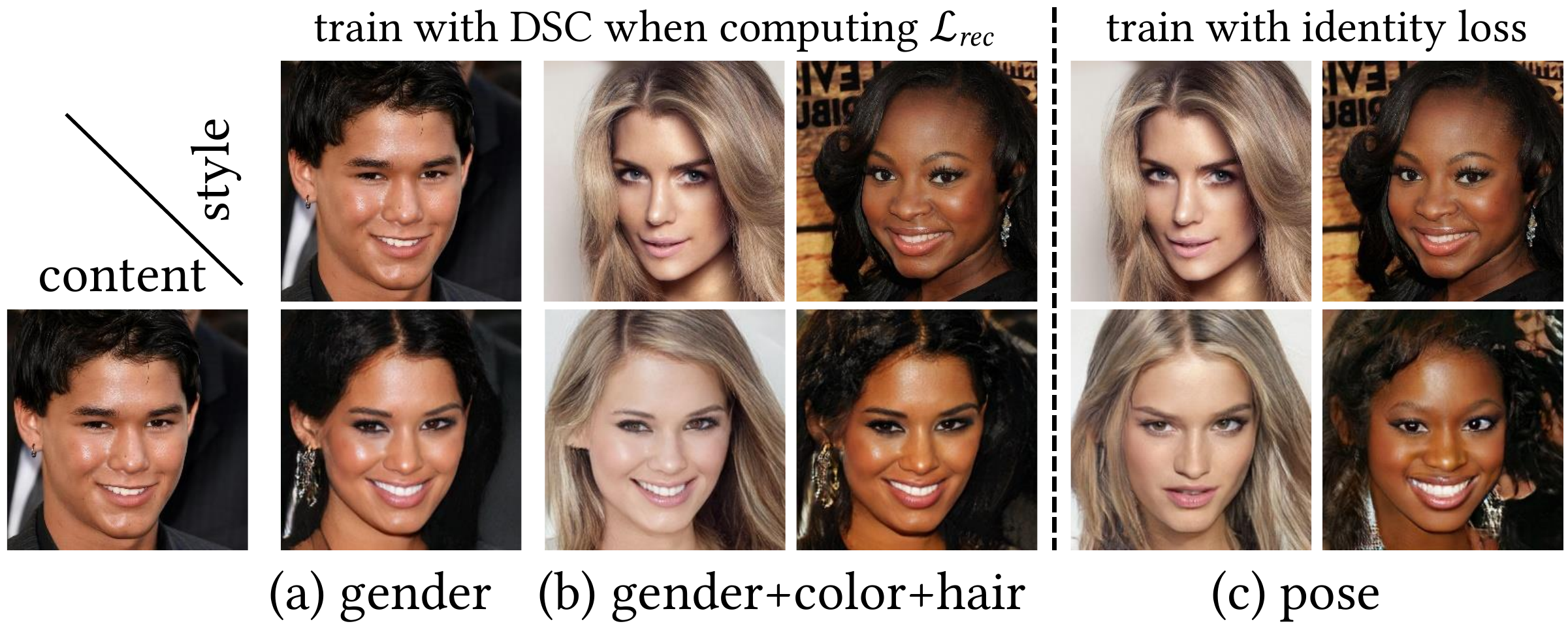}\vspace{-2mm}
\caption{Flexible multi-level attribute transfer.}\vspace{-2mm}
\label{fig:multilevel}
\end{figure}

\noindent
\textbf{Unseen view synthesis.}
Our exemplar-guided framework allows unseen view synthesis. Figure~\ref{fig:interpolation2} shows our synthesized realistic human and cat faces in various pan angles according to the reference faces from the Head Pose Image Database~\cite{gourier2004estimating}.
To transfer identity and prevent low-level content correspondence, we add an identity loss~\cite{deng2019arcface} and do not use DSC for $\mathcal{L}_{rec}$. We further show the usage of DSC for $\mathcal{L}_{rec}$ can flexibly control the facial attribute to be transferred in Fig.~\ref{fig:multilevel}. Using DSC for $\mathcal{L}_{rec}$ helps preserve the identity of the content face, which is suitable for gender and color transfer.
Meanwhile, using identity loss without DSC for $\mathcal{L}_{rec}$, most attributes of the style face except pose can be transferred, which is suitable for pose transfer.

\subsection{Limitations}
\label{sec:limitation}

\begin{figure}[t]
\centering
\includegraphics[width=1\linewidth]{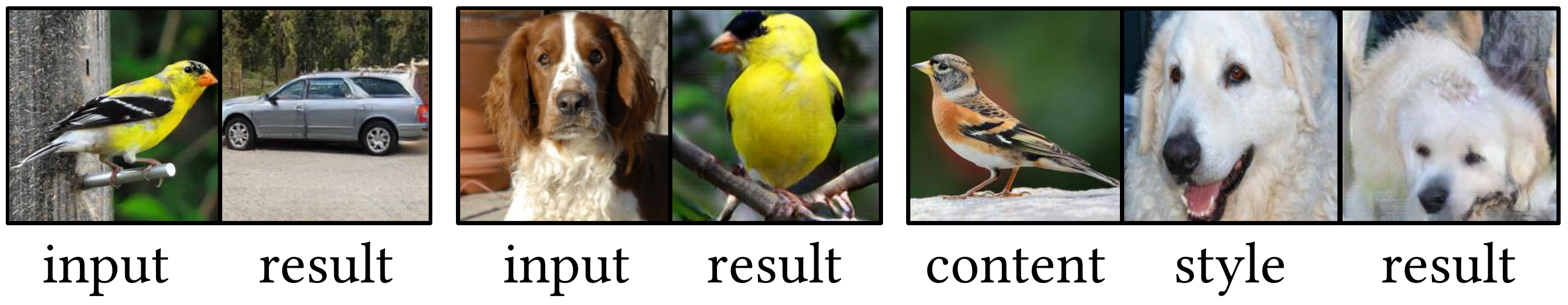}\vspace{-2mm}
\caption{Failure cases caused by (\textit{left}) the lack of semantical supervision, (\textit{middle}) imbalanced training data, and (\textit{right}) scale mismatch of the content and style objects.}\vspace{-4mm}
\label{fig:limitation}
\end{figure}

Figure~\ref{fig:limitation} gives three typical failure cases of our method.
First, it is hard to learn certain semantic correspondences solely from the object appearance, such as which side of a car is its front.
We observe that a bird tail is often translated into the front of a car since they are both the thinner part of the objects.
Second, our method fails to generate a bird sharing the same head direction as the dog, due to the lack of training images of birds looking directly at the camera.
Therefore, special attention should be taken when applying this method to applications where the possible data imbalance issue might lead to biased results towards minority groups  in the dataset.
Finally, when the objects in the content and style images have very different scales, some appearance features cannot be rendered correctly.

\section{Conclusion and Discussion}\vspace{-1mm}

In this paper, we explore the use of GAN generative prior to build a versatile UNIT framework.
We show that the proposed two-stage framework is able to characterize content correspondences at a high semantic level for challenging multi-modal translations between distant domains.
An advantage is that such content correspondences can be discovered with only domain supervision (\ie, only knowing the domain each image belongs to).
We further find in Sec.~\ref{sec:ablation} that fine-level correspondences are learned merely via a generation task.
This might suggest an intriguing behavior of deep neural networks to automatically find and integrate shared appearance features across domains in a coarse-to-fine manner in order to reconstruct various objects. 
It poses a potential of Learning by Generation: building object relationships by generating and transforming them.
Another interesting topic is to learn semantic correspondences beyond object appearance, such as the frontal side of an object discussed in Sec.~\ref{sec:limitation}.
\ys{One possible direction is semi-supervised learning where a small amount of data could be labeled to specify the semantic correspondences.}

\noindent
\ys{\textbf{Acknowledgments.} This study is supported under the RIE2020 Industry Alignment Fund -- Industry Collaboration Projects (IAF-ICP) Funding Initiative, as well as cash and in-kind contribution from the industry partner(s).}

{\small
\bibliographystyle{ieee_fullname}
\bibliography{egbib}
}

\newpage

\appendix

\section{Appendix: Implementation Details}

\subsection{Dataset}

\noindent
\textbf{SynImageNet-291}. For synthesized data, we use the official BigGAN-deep-128 model on TF Hub~\cite{brock2018large} to generate correlated images associated by random latent codes for each of the 291 domains including dogs, wild animals, birds and vehicles. Their class indexes in the original ImageNet 1000 classes are 7$\sim$20, 22, 23, 80$\sim$102, 104, 105, 106, 127, 128, 129, 131$\sim$145, 151$\sim$299, 330$\sim$378, 380, 382$\sim$388, 407, 436, 468, 511, 555, 586, 609, 627, 654, 656, 675, 717, 734, 751, 757, 779, 803, 817, 829, 847, 856, 864, 866, 867, 874.
We apply truncation trick to the latent codes, and obtain 3K images with truncation threshold of $0.5$ and 3K images with truncation threshold of $1.0$.
After filtering low-quality ones, we finally obtain 655 images per domain that are linked across all domains, 600 of which are for training.
For Bird $\leftrightarrow$ Dog or Car, four classes of birds (class index: 10, 11, 12, 13), four classes of dogs (class index: 214, 218, 222, 232) and four classes of cars (class index: 436, 511, 627, 656) are used.

\noindent
\textbf{ImageNet-291}. For each domain $\mathcal{X}$, we first calculate the mean style feature $S_\mathcal{X}$ of the images in $\mathcal{X}$ from \textit{synImageNet-291}. The style feature is defined as the channel-wise mean of the conv5\_2 feature of pre-trained VGG~\cite{SimonyanZ14a}.
Then, we apply HTC~\cite{chen2019hybrid} to ImageNet~\cite{russakovsky2015imagenet} to detect and crop the object regions in the domain $\mathcal{X}$.
Small objects are filtered. The remaining images are ranked based on the similarity between their style features and $S_\mathcal{X}$. We finally select the top 650 images to eliminate outliers, with 600 images for training and 50 images for testing.

\noindent
\textbf{Other datasets}. AFHQ~\cite{choi2020stargan} uses 4K training images and 500 testing images per domain. CelebA-HQ~\cite{karras2018progressive} uses 29K training images and 1K testing images. MS-COCO~\cite{lin2014microsoft} uses 2K giraffe images for training and 197 images for testing. Yosemite~\cite{huang2018multimodal} use 1,231 summer images and 962 winter images for training, and use 309 summer images and 238 winter images for testing.  Metface~\cite{karras2020training} uses 1,336 images for training.

\subsection{Network Architecture}

Let ``C$c$($k$)/$s$'' denote a Convolution-Normalization-Activation layer, with $k\times k$ convolution kernels, output channel number $c$ and stride $s$. If not specified, the default value of $k$ and $s$ are $3$ and $1$, respectively. Let ``FC$c$'' be a fully connection layer with output dimension $c$. ``$\uparrow$'' denotes a $2\times$ nearest upsampling layer. Let ``Res$c$'' denote a Residual Block~\cite{he2016deep} with output channel number $c$ and ``AdaRes$c$'' denote a conditional  ``Res$c$'' with each convolution layer followed by an AdaIN layer.

\noindent
\textbf{Stage I}.
$E_c$:
C64/1-C64/2-C64/1-C128/2-C128/1-C256/2-C256/1-C512/2-C512/1-C512/2-C512/1-C1/1, where ``C'' is a Convolution-IN-LeakyReLU layer.

$E_s$:
C64/2-C128/2-C256/2-C512/2-C512/2-C512/2, where ``C'' is a Convolution-LeakyReLU layer.

$F$:
C512-C512-C512-C512$\uparrow$-C512$\uparrow$-C256$\uparrow$-C128$\uparrow$-C64$\uparrow$-C3, where ``C'' is a Convolution-AdaIN-ReLU layer except the input layer and output layer. The input layer ``C512'' is a Convolution-IN-LeakyReLU layer and the output layer ``C3'' is a Convolution-Tanh layer.
The decoder $F_s$ architecture consists of: \{C512-C512-C512-C512$\uparrow$-C512\}-C3. Layers in $\{\cdot\}$ are shared with $F$ and are conditioned by domain labels, while the remaining layers in $F$ are conditioned by the style features.

$C$:
C32(3)/2-C64(3)/1-C292(4)/1-FC292, where ``C''  is a Convolution-LeakyReLU layer.

\noindent
\textbf{Stage II}.
$E_c$ is the same as in Stage I.

$E_s$:
C64(7)/1-C128(4)/2-C256(4)/2-C256(4)/2-C256(4)/2-C256(4)/2-GAvgPool-FC256-FC64-FC256, where ``C' is a Convolution-ReLU laye. ``GAvgPool'' is a global averaging pooling layer.  ``FC'' is a fully connection layer followed by a ReLU layer.

$G$: AdaRes512-AdaRes512$\uparrow$-AdaRes512-DSC$\uparrow$-AdaRes256-DSC$\uparrow$-AdaRes128$\uparrow$-AdaRes64$\uparrow$-AdaRes64-C3(7), where ``C'' is a Convolution-Tanh layer and ``DSC'' is the proposed dynamic skip connection.

$D$:
C64(3)-Res128$\downarrow$-Res256$\downarrow$-Res512$\downarrow$-Res512$\downarrow$-Res512$\downarrow$-Res512$\downarrow$-C512(4)-C512(1), where ``C''  is a Convolution-IN-LeakyReLU layer. ``$\downarrow$'' is a 2D average pooling layer with kernel size $2\times2$ within the Residual Block for downsampling.

\subsection{Network Training}

\noindent
\textbf{Stage I}.
We adopt the Adam optimizer with a fixed learning rate of 0.0002.
Each iteration uses 16 image pairs from \textit{SynImageNet-291} and 16 images from \textit{ImageNet-291}$+$CelebA-HQ.
We use one NVIDIA Tesla V100 GPU to train our network for 4 epoches (about 44K iterations), which takes about 11 hours.

\noindent
\textbf{Stage II}.
We adopt the Adam optimizer with a fixed learning rate of 0.0001.
The batch size is set to 16.
We use one NVIDIA Tesla V100 GPU to train our network for 75K iterations, which takes about 46 hours.
To compute the style loss, $f_D$ uses the features of the 5th Resblock for Cat $\leftrightarrow$ Human Face, and the 4th Resblock for all other tasks.

\section{Appendix: Supplementary Experiment}

\begin{table*} [t]
\caption{User preference scores in terms of content consistency. Best scores are marked in bold.}\vspace{-1mm}
\label{tb:user_study0}
\centering
\small
\begin{tabular}{l|c|c|c|c|c|c}
\toprule
Task & Male $\leftrightarrow$ Female &  Dog $\leftrightarrow$ Cat & Human Face $\leftrightarrow$ Cat  &  Bird $\leftrightarrow$ Dog &  Bird $\leftrightarrow$ Car &  Average \\
\midrule
TraVeLGAN & 0.017  & 0.015  & 0.004  & 0.015  & 0.008  & 0.012\\
U-GAT-IT & 0.195  & 0.099  & 0.032  & 0.022  & 0.032  & 0.076\\
MUNIT & 0.111  & 0.016  & 0.078  & 0.000  & 0.022  & 0.045 \\
COCO-FUNIT & 0.092  & 0.104  & 0.036  & 0.057  & 0.036  & 0.065\\
StarGAN2 & 0.232  & 0.310  & 0.170  & 0.175  & 0.106 & 0.199\\
\textbf{GP-UNIT} & \textbf{0.353}  & \textbf{0.456}  & \textbf{0.680}  & \textbf{0.731} & \textbf{0.796} & \textbf{0.603}\\
\bottomrule
\end{tabular}
\end{table*}

\begin{table*} [t]
\caption{User preference scores in terms of overall preference. Best scores are marked in bold.}\vspace{-1mm}
\label{tb:user_study1}
\centering
\small
\begin{tabular}{l|c|c|c|c|c|c}
\toprule
Task & Male $\leftrightarrow$ Female &  Dog $\leftrightarrow$ Cat & Human Face $\leftrightarrow$ Cat  &  Bird $\leftrightarrow$ Dog &  Bird $\leftrightarrow$ Car &  Average \\
\midrule
TraVeLGAN & 0.006  & 0.012  & 0.000  & 0.000  & 0.009  & 0.006 \\
U-GAT-IT & 0.162  & 0.079  & 0.001  & 0.004  & 0.005  & 0.050 \\
MUNIT & 0.099  & 0.007  & 0.053  & 0.000  & 0.009  & 0.033 \\
COCO-FUNIT & 0.098  & 0.085  & 0.000  & 0.033  & 0.004  & 0.044 \\
StarGAN2 & 0.240  & 0.240  & 0.153  & 0.157  & 0.063  & 0.171 \\
\textbf{GP-UNIT} & \textbf{0.394}  & \textbf{0.576}  & \textbf{0.793}  & \textbf{0.805}  & \textbf{0.910}  & \textbf{0.696} \\
\bottomrule
\end{tabular}
\end{table*}

\subsection{User Study}

We conduct a user study to evaluate the input-output content consistency and overall translation performance. A total of 25 subjects participate in this study to select the best ones from the results of six methods. Because in some tasks like Male $\leftrightarrow$ Female, the performance of each method is similar, we allow multiple selections. For each selection, if a user select results from $N$ methods as the best results, those methods get $1/N$ scores, and other methods get $0$ scores. A total of 2,500 selections on 50 groups of results (\ie, Figs.~\ref{fig:comparison00}-\ref{fig:comparison04}) are tallied. Table~\ref{tb:user_study0} and Table~\ref{tb:user_study1} demonstrate the average user scores, where the proposed method receives notable preference for both content consistency and overall performance.

\begin{figure}[htbp]
\centering
\includegraphics[width=1\linewidth]{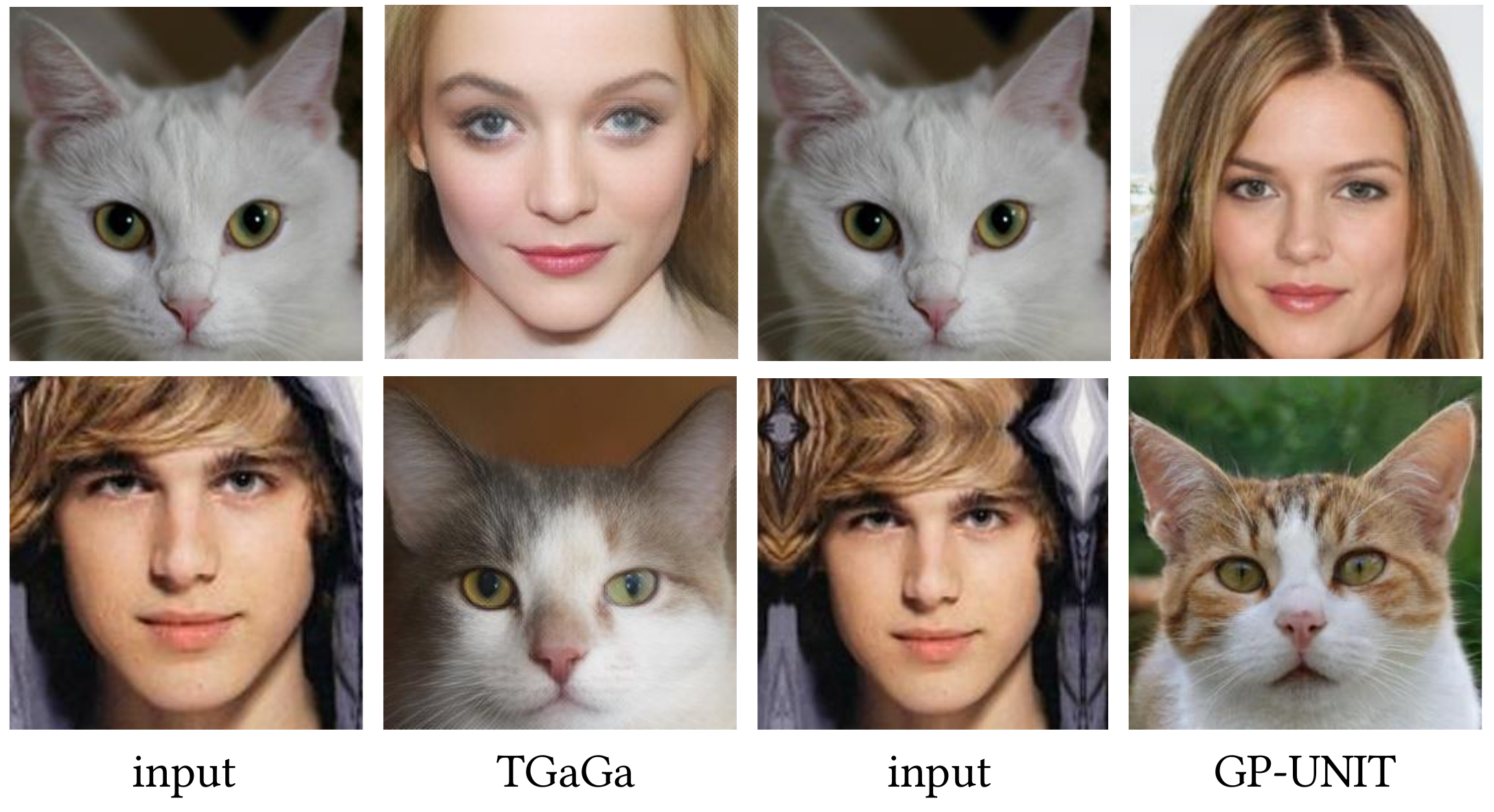}\vspace{-2mm}
\caption{Comparison in geometry-preserving with TGaGa.}
\label{fig:tgaga1}\vspace{-3mm}
\end{figure}

\subsection{Comparison with TGaGa}
\label{sec:tgaga}

\begin{figure*}[htbp]
\centering
\includegraphics[width=0.8\linewidth]{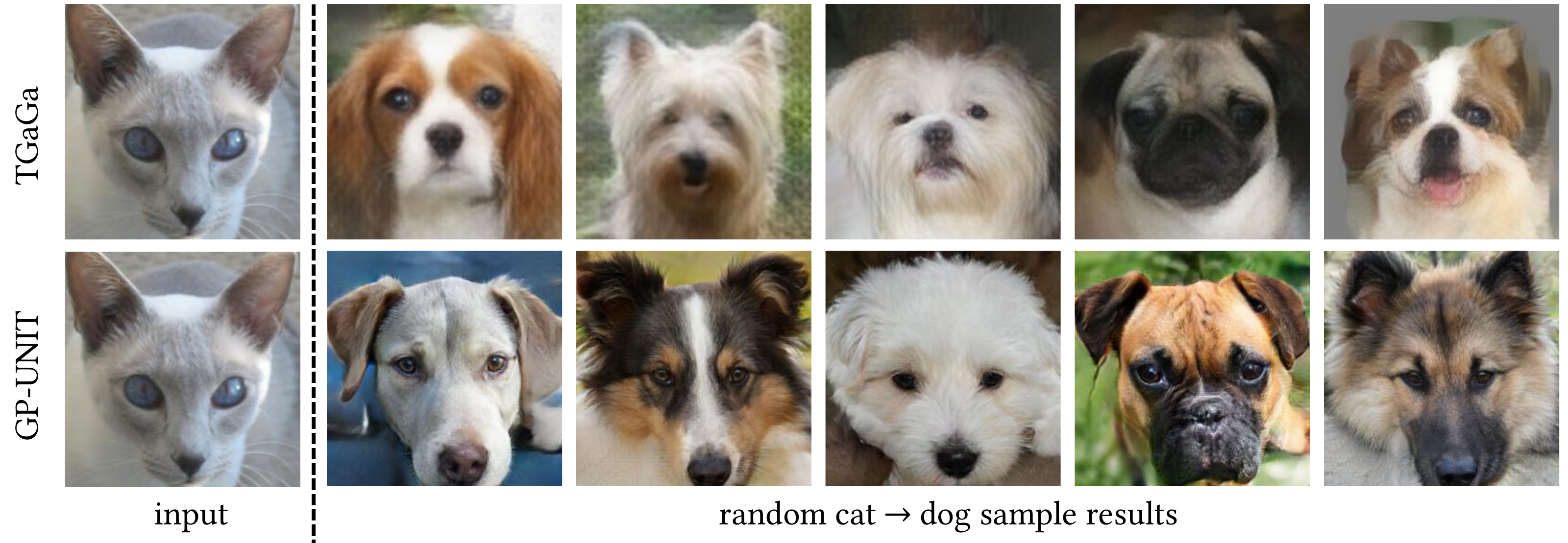}\vspace{-2mm}
\caption{Comparison in multi-modal generation with TGaGa.}\vspace{-2mm}
\label{fig:tgaga2}
\end{figure*}

In Fig.~\ref{fig:tgaga1}, we present translation results with large geometric deformations. As one of the most related works to ours that deal with drastic geometric deformations, TGaGa~\cite{wu2019transgaga} and our method both effectively build geometric correspondences between two distant domains. The main superiority of GP-UNIT over TGaGa lies in more realistic texture generation. We compare with TGaGa on multi-modal generation in Fig.~\ref{fig:tgaga2}. The results of TGaGa are blurry and our method generates more vivid details.

For large geometric deformations, there is an inherent content-style trade-off problem:
\begin{itemize}[itemsep=1.5pt,topsep=1pt,parsep=0pt]
  \item \textbf{Content-style trade-off}: Due to the inherent differences in the proportions of facial features of different species, it is impossible to generate a realistic human or dog face from a cat face with the locations of eyes/nose/mouth unchanged. A robust method should adjust the locations of such facial features to match the style of target domains while maintaining the original geometry as much as possible. Therefore, there is a trade-off between realism and content consistency.
  \item \textbf{Cycle consistency overemphasizes content}. Standard cycle consistency is often too restrictive and results in only texture transfer without geometry adjustment. Therefore, methods like MUNIT overemphasize content consistency, sacrificing realism as in Fig.~\ref{fig:comparison}.
  \item \textbf{GP-UNIT strikes a good balance}. TGaGa solves this problem with explicit geometry adjustment, while we learn a high-level correspondence based on which only necessary mid-level correspondences are then added. It can be seen in Figs.~\ref{fig:tgaga1}-\ref{fig:tgaga2} that both TGaGa and GP-UNT successfully adjust the geometry, but in our results, the locations of the facial features better match the input (although not exactly the same), which proves that GP-UNIT strikes a better balance between realism and content consistency than TGaGa.
\end{itemize}

\begin{figure*}[htbp]
\centering
\includegraphics[width=0.92\linewidth]{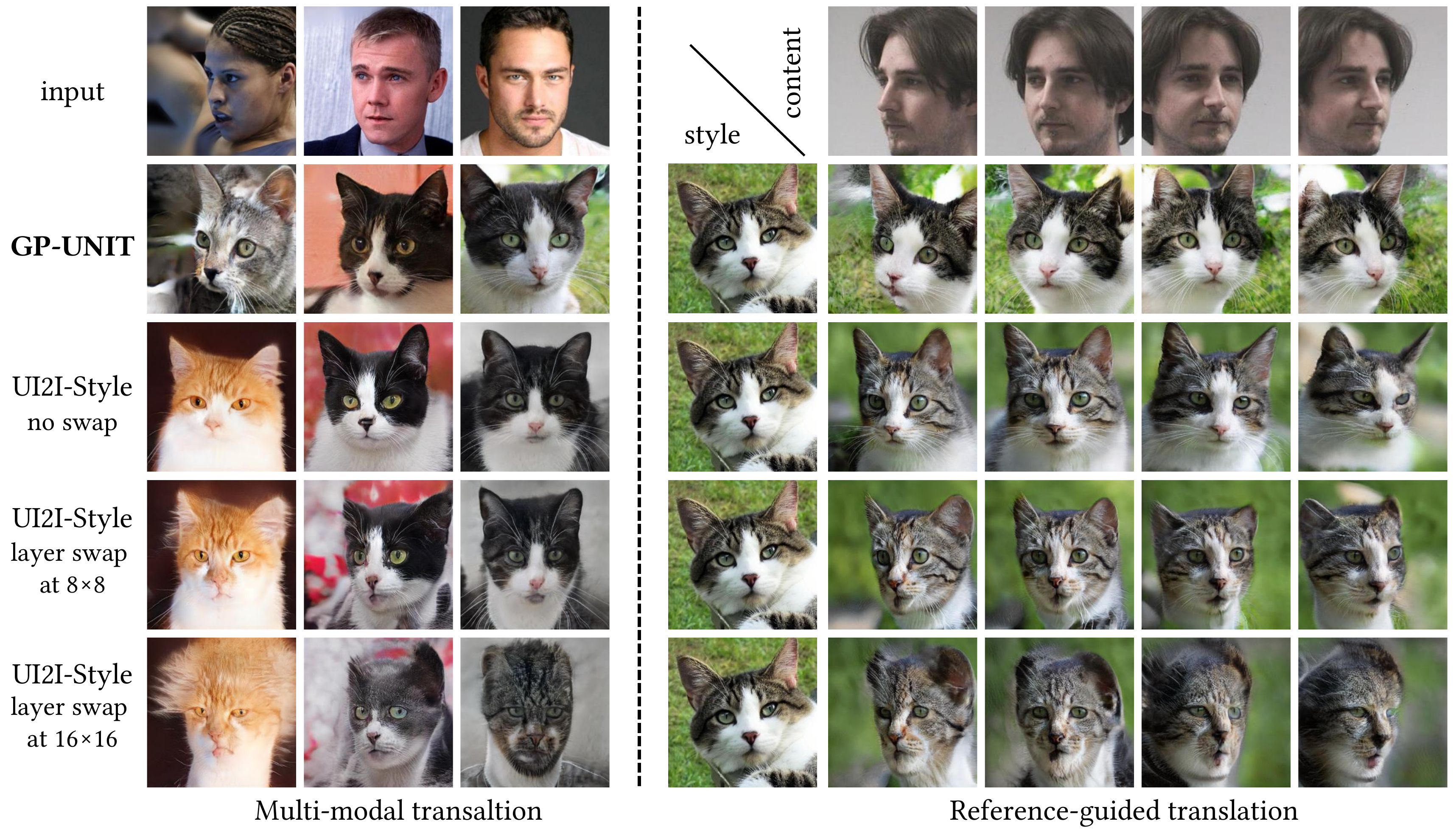}\vspace{-2mm}
\caption{Visual comparison on Human Face $\rightarrow$ Cat with UI2I-Style~\cite{kwong2021unsupervised}.}\vspace{-2mm}
\label{fig:stylegan1}
\end{figure*}

\subsection{Comparison with UI2I-Style}

StyleGAN-based methods achieves unsupervised translations between two domains using finetuning~\cite{pinkney2020resolution,kwong2021unsupervised}\footnote{Although the W+ space of StyleGAN enables image reconstruction in arbitrary domains~\cite{abdal2019image2stylegan}, the W+ latent code still does not support semantic editing like interpolation, style fusing and translation between domains beyond the domains StyleGAN is trained or finetuned on.}. By assuming a small distance between the models before and after finetuning,  images generated by the original StyleGAN and the finetuned StyleGAN belong to two domains, respectively, but have strong content correspondences. Layer-swap is proposed by~\cite{pinkney2020resolution} to control how many content features from the source domain are preserved. UI2I-Style~\cite{kwong2021unsupervised} shows good results on Face $\rightarrow$  Art Portrait and Cat $\rightarrow$ Dog with layer-swap at resolution $16\times 16$. However, when we apply UI2I-Style to domains with more visual discrepancies like human face and cat, layer-swap results in fused and very unreal results as shown in Fig.~\ref{fig:stylegan1}. Even without using layer-swap, the assumption of a small distance between the models does not hold. Thus, the content correspondences using the same latent code are drastically weakened. For example, the position of the eyes of the cats does not match those of human faces. By comparison, the results of our method are better in content consistency.

\begin{table*} [htbp]
\caption{Ablation study on DIF, Diversity and input-output content consistency. Best scores are marked in bold.}\vspace{-1mm}
\label{tb:user_study2}
\centering
\small
\begin{tabular}{l|c|c|c}
\toprule
Metric & FID & Diversity & Content Consistency \\
\midrule
without generative prior	& 16.11	& \textbf{0.55}	& 0.02 \\
without dynamic skip connection	& 15.83	& 0.52	& 0.15 \\
full model	& \textbf{15.29}	& 0.51	& \textbf{0.83} \\
\bottomrule
\end{tabular}
\end{table*}

\subsection{Quantitative Evaluation}

\noindent
\textbf{Ablation study}. To better understand the effect of the distilled generative prior and dynamic skip connection, we perform quantitative comparison in terms of quality, diversity and content consistency.
For content consistency, ten users are invited to select the best one from the results of three configurations in terms of content consistency. FID, Diversity averaged over the whole testing set and the user score averaged over six groups of results are presented in Table~\ref{tb:user_study2}.
\begin{itemize}[itemsep=1.5pt,topsep=1pt,parsep=0pt]
  \item \textbf{FID}: Results of our full model have better quality.
  \item \textbf{Diversity}: Three configurations have comparable diversities. With fewer content constraints, results of model without the generative prior or without dynamic skip connection are more diverse.
  \item \textbf{Content consistency}: The generative prior and the dynamic skip connection effectively help our model better capture high-level and mid-level content correspondences (high Content Consistency).
\end{itemize}

\noindent
\ys{\textbf{Discussion on content correspondence}. In this paper, we mainly use user scores to evaluate the content consistency.
For objective evaluation, landmark correspondence might be one potential metric. We conduct human/cat face landmark detection to predict eye and nose correspondences. GP-UNIT is comparable to other baselines in normalized point-to-point error (GP-UNIT/MUNIT/StarGAN2/COCO-FUNIT/TraVeLGAN:~0.27/0.23/0.19/0.36/0.31).
However, these scores still does not well match the subjective user scores in Table~\ref{tb:user_study}.
The reason is the \textbf{content-style trade-off problem} as discussed in Sec.~\ref{sec:tgaga}. A robust method should adjust the locations of facial features to match the target domains, which does not favor this metric. Since this adjustment is task-dependent, it is nearly impractical to define a universal metric.
Therefore, landmark correspondence is less explored as evaluation metrics in UNIT baselines. UNIT still lacks a good objective evaluation metric for content consistency, which is an important research direction.}

\subsection{Style Blending}

\begin{figure}[htbp]
\centering
    \includegraphics[width=1\linewidth]{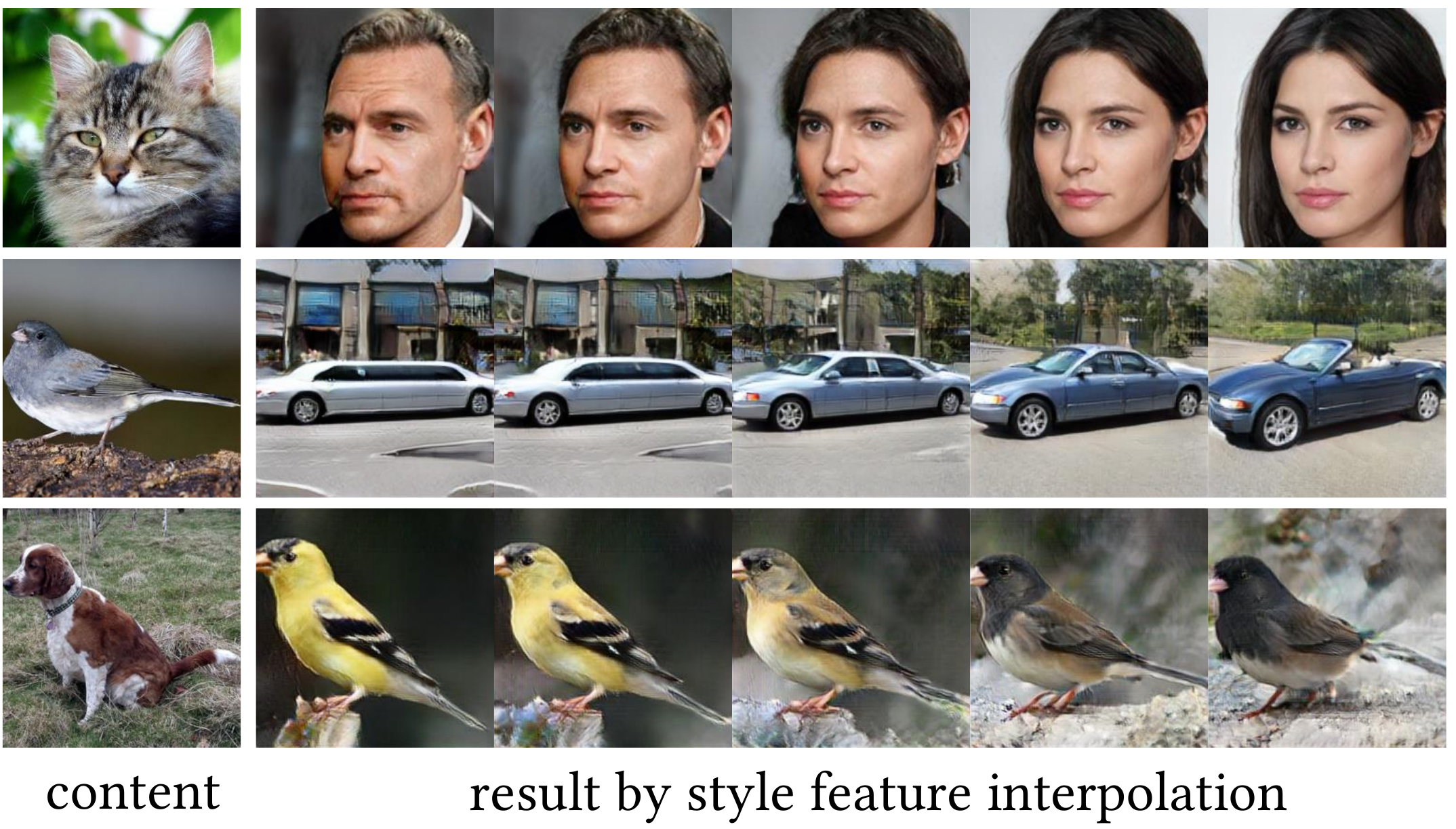}\vspace{-2mm}
\caption{Style blending.}\vspace{-1mm}
\label{fig:interpolation1}
\end{figure}

As shown in Fig.~\ref{fig:interpolation1}, we perform a linear interpolation to style feature, and observe smooth changes along with the latent space from one to another.  It implies a reasonable coverage of the manifold captured by our method, which can be used for novel unseen style rendering.

\subsection{Visual Comparison with the State of the Arts}

\begin{figure*}[htbp]
\centering
\includegraphics[width=1\linewidth]{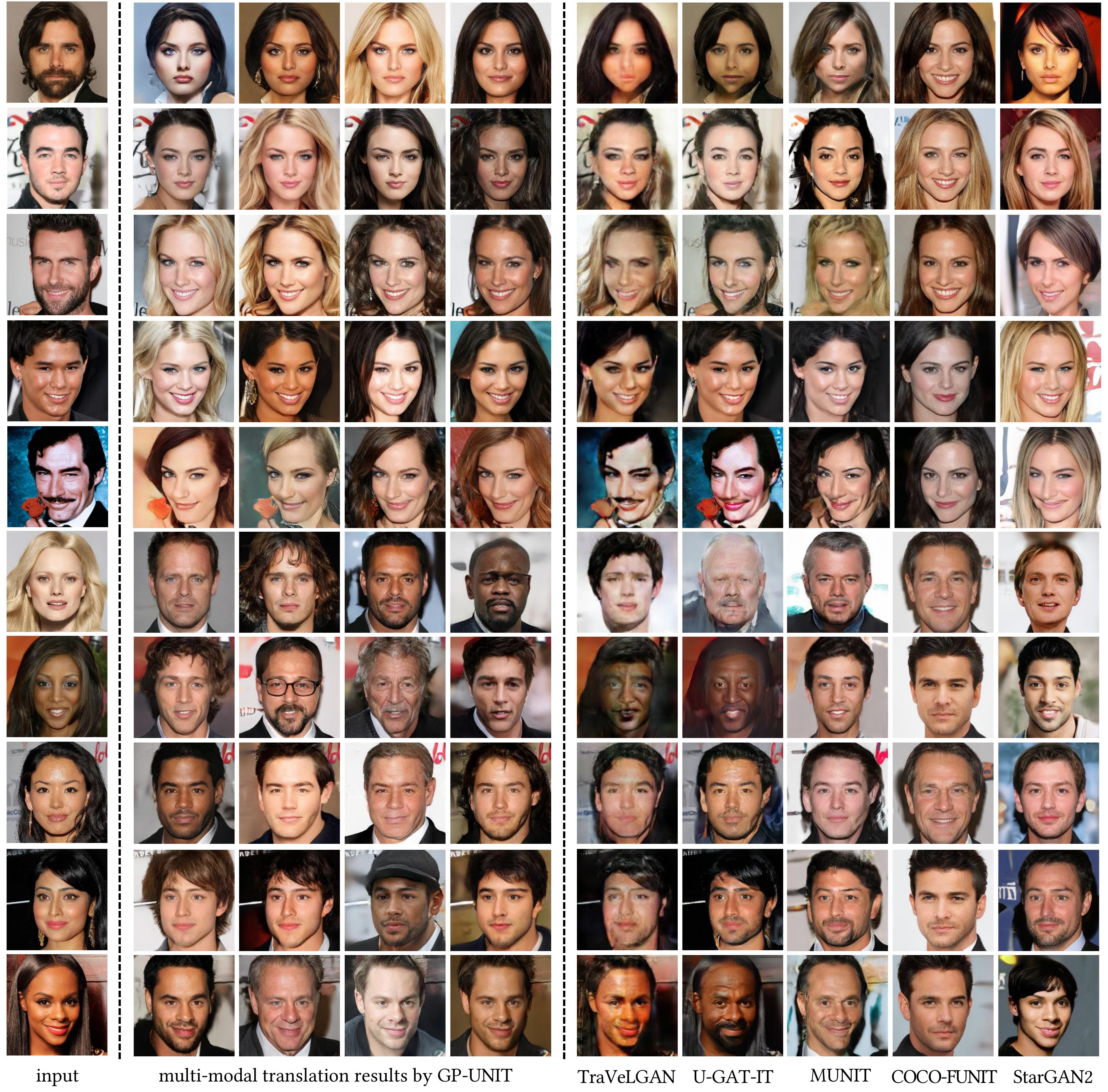}
\caption{Comparison on Male$\leftrightarrow$Female with TraVeLGAN~\cite{amodio2019travelgan}, U-GAT-IT~\cite{kim2019u}, MUNIT~\cite{huang2018multimodal}, COCO-FUNIT~\cite{saito2020coco} and StarGAN2~\cite{choi2020stargan}.}
\label{fig:comparison00}
\end{figure*}

\begin{figure*}[htbp]
\centering
\includegraphics[width=1\linewidth]{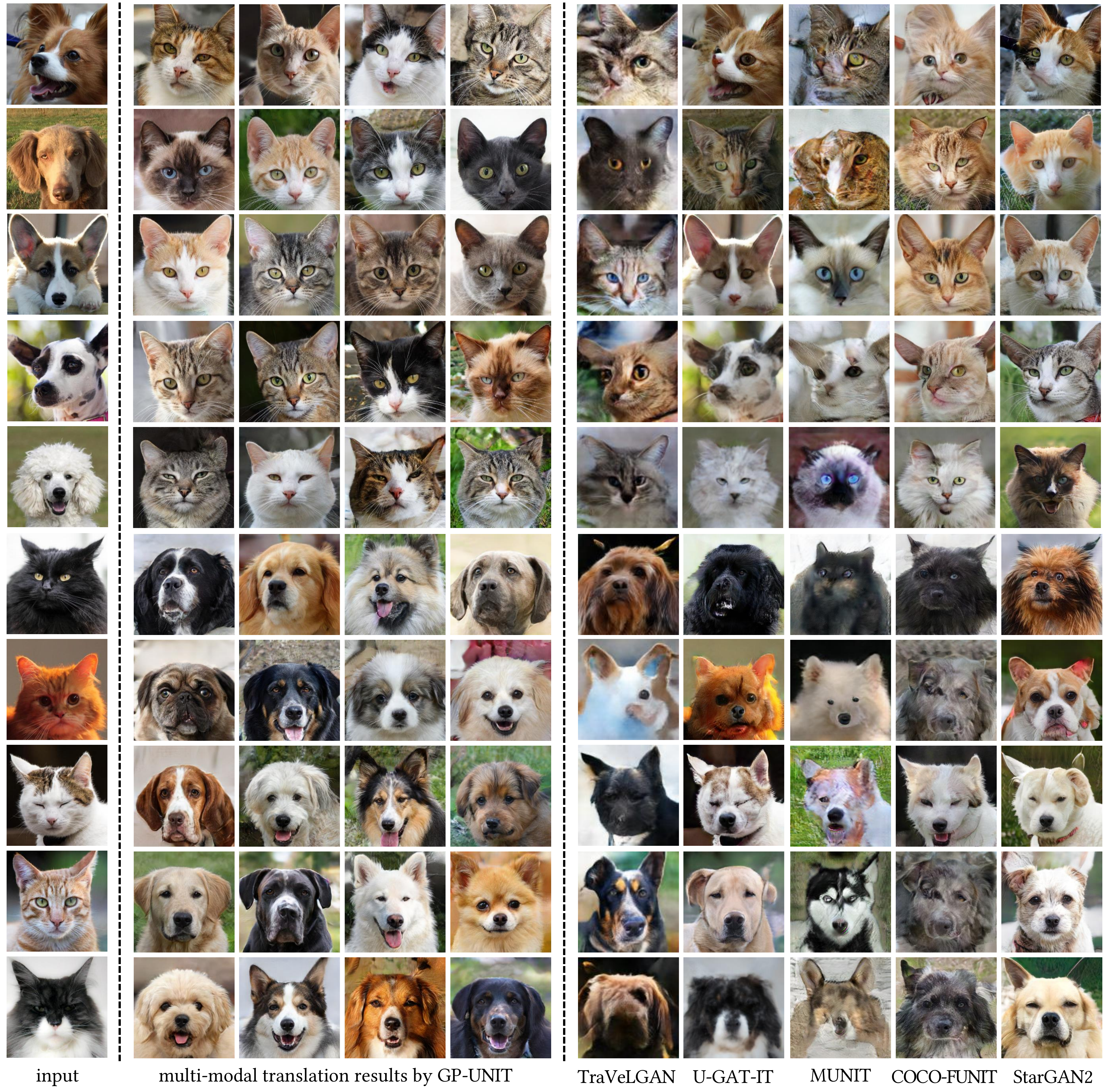}
\caption{Comparison on Dog$\leftrightarrow$Car with TraVeLGAN~\cite{amodio2019travelgan}, U-GAT-IT~\cite{kim2019u}, MUNIT~\cite{huang2018multimodal}, COCO-FUNIT~\cite{saito2020coco} and StarGAN2~\cite{choi2020stargan}.}
\label{fig:comparison01}
\end{figure*}

\begin{figure*}[htbp]
\centering
\includegraphics[width=1\linewidth]{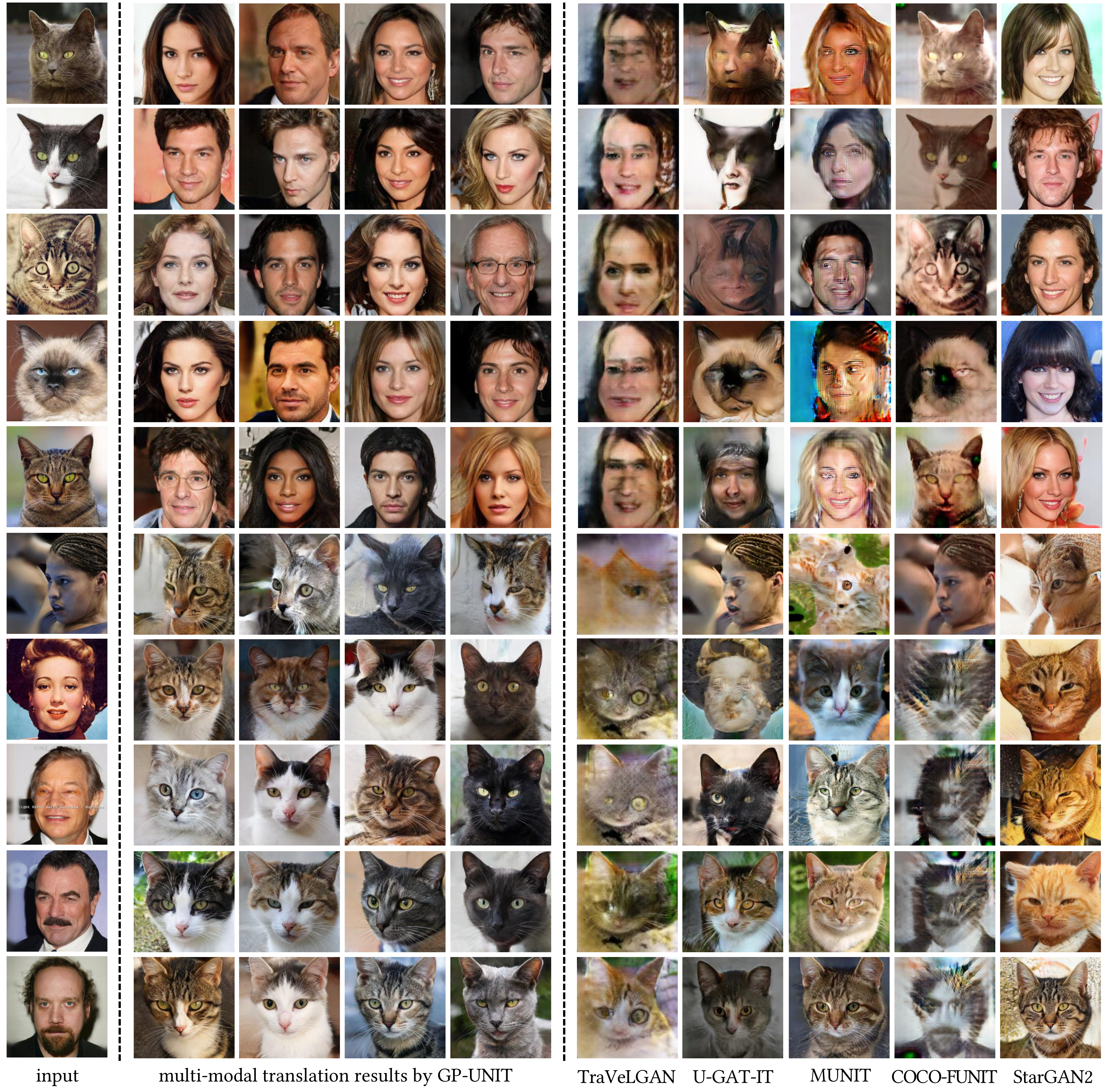}
\caption{Comparison on Cat$\leftrightarrow$Human with TraVeLGAN~\cite{amodio2019travelgan}, U-GAT-IT~\cite{kim2019u}, MUNIT~\cite{huang2018multimodal}, COCO-FUNIT~\cite{saito2020coco} and StarGAN2~\cite{choi2020stargan}.}
\label{fig:comparison02}
\end{figure*}

\begin{figure*}[htbp]
\centering
\includegraphics[width=1\linewidth]{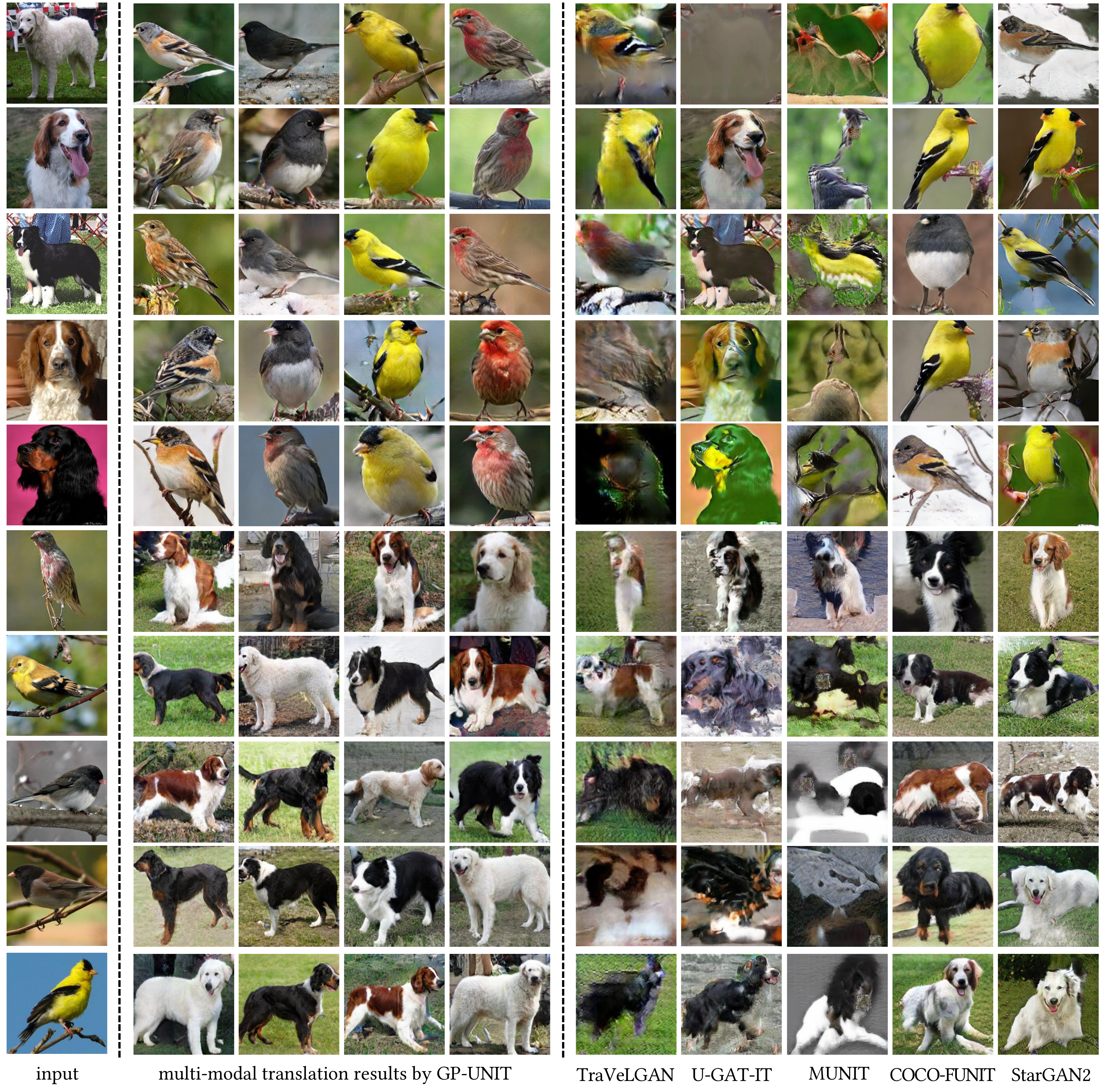}
\caption{Comparison on Dog$\leftrightarrow$Bird with TraVeLGAN~\cite{amodio2019travelgan}, U-GAT-IT~\cite{kim2019u}, MUNIT~\cite{huang2018multimodal}, COCO-FUNIT~\cite{saito2020coco} and StarGAN2~\cite{choi2020stargan}.}
\label{fig:comparison03}
\end{figure*}

\begin{figure*}[htbp]
\centering
\includegraphics[width=1\linewidth]{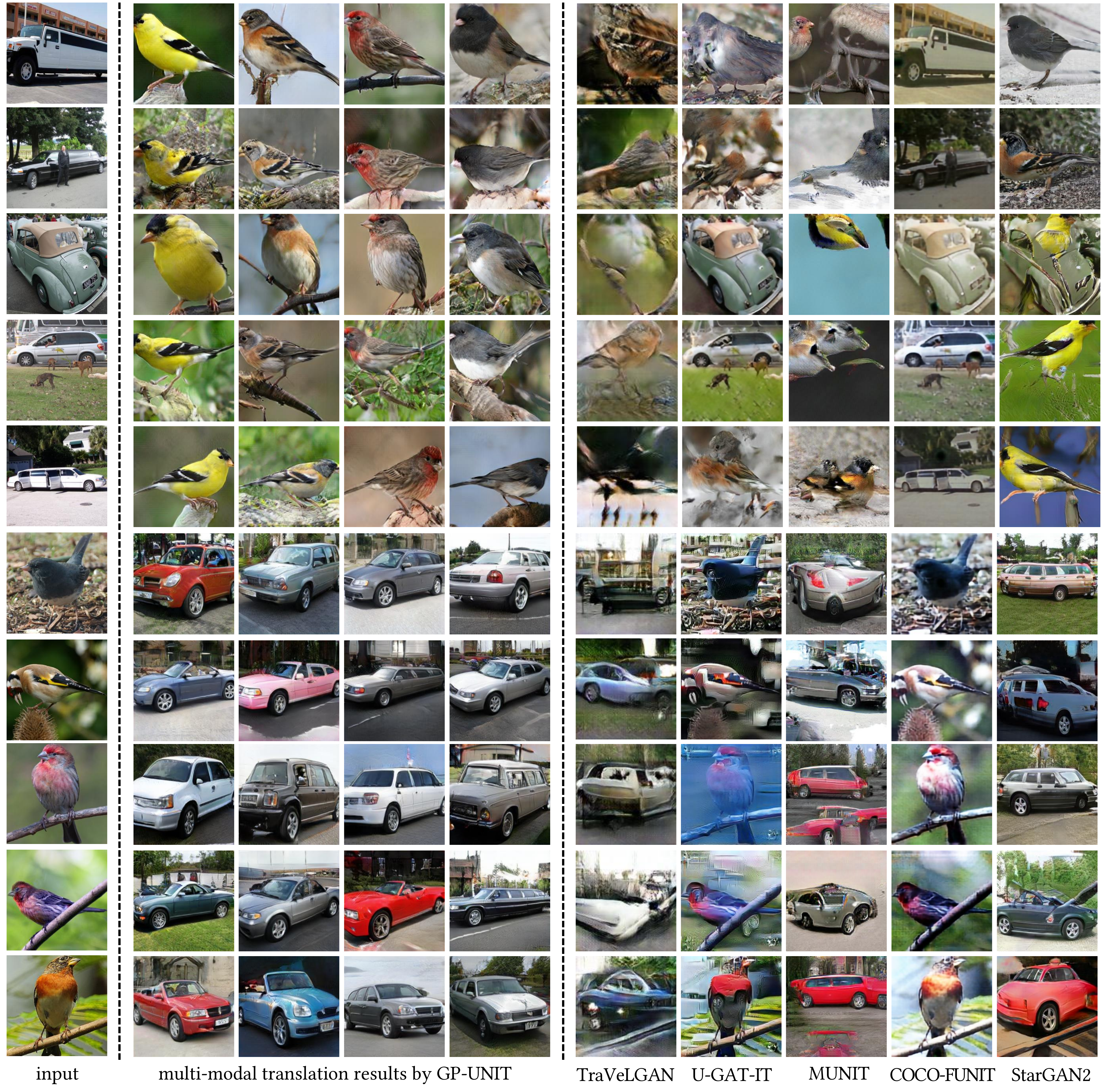}
\caption{Comparison on Car$\leftrightarrow$Bird with TraVeLGAN~\cite{amodio2019travelgan}, U-GAT-IT~\cite{kim2019u}, MUNIT~\cite{huang2018multimodal}, COCO-FUNIT~\cite{saito2020coco} and StarGAN2~\cite{choi2020stargan}.}
\label{fig:comparison04}
\end{figure*}

In addition to the examples shown in the main paper,
we show more visual comparison results with TraVeLGAN~\cite{amodio2019travelgan}, U-GAT-IT~\cite{kim2019u}, MUNIT~\cite{huang2018multimodal}, COCO-FUNIT~\cite{saito2020coco} and StarGAN2~\cite{choi2020stargan} in Figs.~\ref{fig:comparison00}-\ref{fig:comparison04}.
Our method surpasses these methods in:
\begin{itemize}[itemsep=1.5pt,topsep=1pt,parsep=0pt]
  \item more accurate content correspondences with the input images;
  \item less artifacts caused by the domain-specific information leakage form the input images;
  \item better matched shape and appearance features with the target domain;
  \item more realistic image details.
\end{itemize}

\end{document}